\documentclass[lettersize,journal]{IEEEtran}
\usepackage{amsmath,amsfonts}
\usepackage{array}
\usepackage[caption=false,font=normalsize,labelfont=sf,textfont=sf]{subfig}
\usepackage{textcomp}
\usepackage{stfloats}
\usepackage{url}
\usepackage{verbatim}
\usepackage{graphicx}
\usepackage{cite}
\usepackage{epsfig}
\usepackage{amssymb}
\usepackage{multirow}
\usepackage{algorithm}
\usepackage{algpseudocode}
\usepackage{hyperref}
\usepackage{enumitem}
\usepackage{pifont}
\usepackage{xcolor}

\usepackage{wrapfig,lipsum,booktabs}

\begin{document}

\title{MD-ProjTex: Texturing 3D Shapes with Multi-Diffusion Projection}

\author{
    Ahmet Burak Yildirim, Mustafa Utku Aydogdu, Duygu Ceylan, Aysegul Dundar
    \IEEEcompsocitemizethanks{
        \IEEEcompsocthanksitem A. B. Yildirim, M. U. Aydogdu, A. Dundar are with Department of Computer Science, Bilkent University, Ankara, Turkey
        \IEEEcompsocthanksitem  D. Ceylan is with Adobe.
    }
}

\maketitle

\begin{abstract}
We introduce MD-ProjTex, a method for fast and consistent text-guided texture generation for 3D shapes using pretrained text-to-image diffusion models.  
At the core of our approach is a multi-view consistency mechanism in UV space, which ensures coherent textures across different viewpoints. Specifically, MD-ProjTex fuses noise predictions from multiple views at each diffusion step and jointly updates the per-view denoising directions to maintain 3D consistency.  
In contrast to existing state-of-the-art methods that rely on optimization or sequential view synthesis, MD-ProjTex is computationally more efficient and achieves better quantitative and qualitative results. 
Please refer to the web page of the project for 3D visuals: \href{https://mdprojtex.abyildirim.com/}{https://mdprojtex.abyildirim.com/}
\end{abstract}


\section{Introduction}

The creation of 3D content holds fundamental importance across various domains, including virtual reality (VR), augmented reality (AR), gaming, robotics, and simulations. While substantial advancements have been made in 2D image generation through GANs \cite{karras2020analyzing} and diffusion models \cite{saharia2022photorealistic}, attention has now shifted towards the next significant challenge: high-quality 3D content generation. Unlike 2D image generation, achieving controllability over viewpoints is crucial in 3D content generation endeavors.

Besides methodological advancements, the availability of vast image datasets plays a crucial role in the success of 2D image generation. For instance, ImageGen \cite{saharia2022photorealistic} is trained on 860 million image-text pairs, while Stable diffusion v2 \cite{rombach2022high} utilizes a dataset comprising 5.85 billion image-text pairs. In contrast, the collection of 3D data is often prohibitively expensive, with efforts predominantly focused on specific objects like faces \cite{gecer2019ganfit} and human bodies \cite{zhang2019predicting,lattas2020avatarme}. Consequently, researchers have proposed methods to infer 3D content from either 2D images \cite{chan2022efficient,or2022stylesdf,dundar2023fine,dundar2023progressive} or 2D image generator models \cite{lin2023magic3d}. Recently, it has been demonstrated that 2D text-to-image generators trained on extensive image corpora can effectively guide the process of generating textures for 3D geometry \cite{mohammad2022clip,richardson2023texture,chen2023text2tex}.

\begin{figure*}[h!]
\centering
\includegraphics[width=\textwidth]{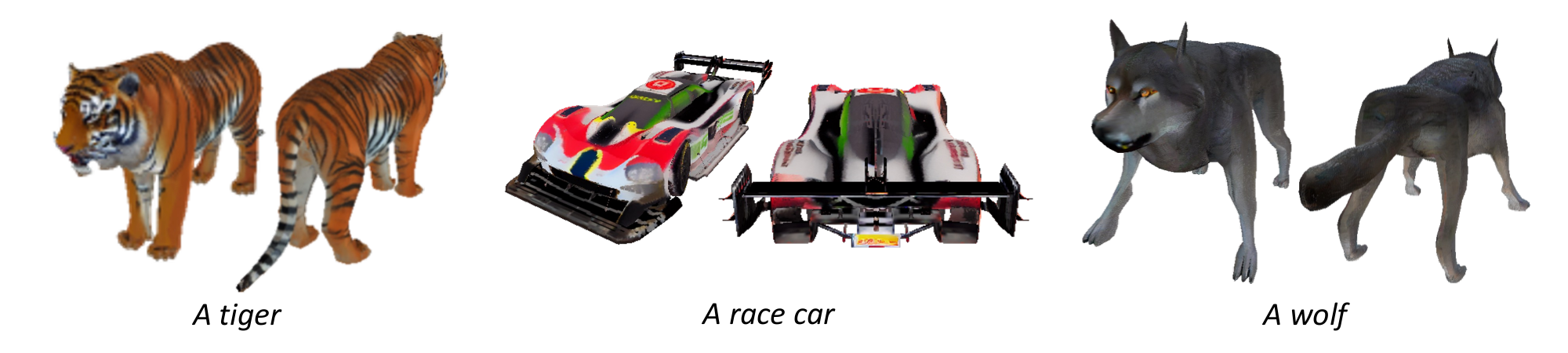}
\caption{Our framework can texture 3D models based on given prompts. 
It is training-free and fast due to the parallel generation of multiple views with the multi-diffusion approach applied to the projected textures.}
\label{fig:teaser}
\end{figure*}

State-of-the-art models that rely on  2D text-to-image diffusion models for texturing of 3D shapes, TEXTure \cite{richardson2023texture} and Text2Tex \cite{chen2023text2tex}, approach texture generation as an inpainting task. Beginning from a random viewpoint, they initially generate an image using depth-conditioned diffusion models. This generated image serves as the foundation for texture projection. To complete the texture, the object undergoes rendering from various viewpoints, with void regions in the texture filled via diffusion models in image space. Despite their promising outcomes, these methods exhibit slowness due to sequential image generation conditioned on progressively advancing UV maps. Moreover, they often struggle to maintain consistency in texture generation, as style discrepancies accumulate during the gradual texture-filling process based on image space inpainting.

On the other hand, various methods have been proposed to train or fine-tune base models for 3D generation tasks. For example, Meta 3D TextureGen \cite{bensadoun2024meta} adapts the Stable Diffusion (SD) model by combining position and normal maps, a process that requires additional training.  The DreamMat model \cite{zhang2024dreammat} aims to generate physically-based rendering materials by fine-tuning a light-aware 2D diffusion model, while Paint3D \cite{zeng2024paint3d} trains a diffusion model specifically for UV inpainting. Unlike these approaches, which necessitate task-specific training, our method addresses texture generation without requiring any additional training.

In this paper, we present \textbf{MD-ProjTex}, which uses multi-diffusion technique \cite{bar2023multidiffusion} in the uv-space. 
During each denoising step, the multi-diffusion algorithm integrates denoising directions obtained from the reference model across different views in a common texture map defined in the uv-space. It endeavors to adhere closely to these directions while considering the shared pixels among different views. 
Because our algorithm provides consistency during the generation, it is significantly faster than methods that generate images from different views sequentially.

In summary, our main contributions are as follows:
\begin{itemize}[leftmargin=*]
    \item We propose a novel pipeline that achieves text-prompted texture generation via multi-diffusion-based texture projection.
    Prompt-based results are shown in Fig. \ref{fig:teaser}.
    \item Our method is training-free and does not require run-time optimization. Additionally, we denoise multiple views in parallel while generating the textures. It does not rely on sequential generation, inpainting, or re-generation methods, nor does it require training a new model. 
    Because the multi-view consistency is merged into our parallel generation pipeline, our method is significantly faster than the competing methods. 
    \item We conduct extensive experiments to show the effectiveness of our framework. Quantitative and qualitative results show the superiority of our method compared to state-of-the-art.

\end{itemize}

\section{Related Work}

\textbf{3D Generation Models.}
Training 3D generation models on 3D data has been achieved only for a select few objects, such as faces \cite{gecer2019ganfit} and human bodies \cite{zhang2019predicting,lattas2020avatarme,altindis2023refining}. This is primarily due to the expensive equipment and laboratory setups required for collecting 3D object data, which also limits the diversity of images.
To extend 3D generation and reconstruction to arbitrary objects, researchers have shifted their focus to learning 3D attributes from images. Building upon the success of StyleGAN models \cite{karras2020analyzing} in 2D generations and NERF-based learning for 3D models \cite{mildenhall2020nerf}, 3D-aware generative models have been proposed. These models combine adversarial learning with differentiable rendering \cite{nguyen2019hologan,niemeyer2021giraffe,chan2021pi,gu2021stylenerf,gao2022get3d,chan2022efficient}. They incorporate a 3D-aware structure in their networks, and through differentiable rendering, these 3D-aware representations are projected into 2D images. Discriminators are trained on sets of real 2D images and those generated by the generator.
Despite achieving successful generations, these models have been limited to a few objects. One limitation is that different networks are trained for each domain, and each training requires a curated dataset for each object, such as the FFHQ dataset \cite{karras2020analyzing} for faces and the LSUN dataset \cite{yu2015lsun} for cars and horses. Another limitation is that they do not provide strict multi-view consistency because the 3D-aware structure is followed by convolution-based upsampling layers to achieve high-resolution image outputs.

\noindent \textbf{Text-to-Image Diffusion Models.} 
Text-to-image diffusion models have emerged as a significant breakthrough in deep learning-based image generation over the past three years. Fueled by large-scale text-image paired datasets, these models have delivered impressive results in text-conditioned image synthesis \cite{nichol2021glide,saharia2022photorealistic,ramesh2022hierarchical,avrahami2023spatext,balaji2022ediffi}. Given a text prompt, these models iteratively denoise to produce realistic images.
Many of these works have made their models open-source, including powerful Stable-diffusion models \cite{rombach2022high}, which have facilitated extensive research leveraging pretrained models \cite{avrahami2022blended,chen2023text2tex,richardson2023texture}. These models have also been extended to accept conditional inputs other than prompts, such as depth maps and edges, through models like ControlNet \cite{zhang2023adding}.
In our work, we employ a pretrained Stable-diffusion model \cite{rombach2022high} with depth-conditioning and line-art conditioning \cite{zhang2023adding}. Our approach requires no training and is capable of texturing 3D shapes via multi-diffusion projection onto the textures.

\noindent \textbf{Text-guided 3D Generation Methods.}
For the text-guided 3D generation of objects, score distillation loss is proposed \cite{poole2022dreamfusion}, which optimizes a NERF for a given prompt with the gradients coming from the diffusion models.
This loss is used by different methods to guide 3D generation models optimized for each generation 
\cite{lin2023magic3d,metzer2023latent}. 
However, this run-time optimization of NERF models is very costly.
Recently, TEXTure \cite{richardson2023texture}, Text2Tex \cite{chen2023text2tex}, and TexGEN \cite{huo2024texgen} have demonstrated improved runtime efficiency compared to previous NERF-based optimization methods. These approaches frame texture generation as an inpainting problem. They initiate the process by generating an initial image from a random viewpoint using depth-conditioned diffusion models. This initial image is then projected onto a texture space. To complete the texture, the object undergoes rendering from various perspectives, and the regions lacking texture are filled in the image space using diffusion models. Despite the promising outcomes of these methods, they suffer from sluggishness due to the sequential generation of images, relying on gradually advancing UV maps. Moreover, they often encounter challenges in maintaining consistency in texture generation, as the incremental filling of texture based on image space inpainting tends to accumulate slight style discrepancies over time. Finally, they are also slow due to the sequential generation of images via multiple denoising steps. 
Other texture generation methods, such as Paint-it \cite{youwang2024paint}, also involve lengthy computation times due to their optimization of the U-Net model. Similarly, TexPainter \cite{zhang2024texpainter} proposes a denoising process from multiple views, blending them into a common color-space texture image. This approach optimizes the latent space of the decoder through backpropagation, which further contributes to extended computational times.

In a different line of research, several methods have been proposed to train base models for 3D generation tasks. For instance, Meta 3D TextureGen \cite{bensadoun2024meta} fine-tunes the Stable Diffusion (SD) model by concatenating position and normal maps, which requires additional training. TextureDreamer \cite{yeh2024texturedreamer} transfers textures from a small set of input images (3 to 5) to target 3D shapes across various categories, relying on a few reference style images and DreamBooth fine-tuning. The DreamMat model \cite{zhang2024dreammat} focuses on generating physically-based rendering materials by fine-tuning a light-aware 2D diffusion model, while Paint3D \cite{zeng2024paint3d} aims to train a diffusion model for UV inpainting. In contrast to these methods, which require task-specific training, our approach seeks to address the texture generation task without the need for any additional training.

Finally, there are other works such as PointUV \cite{yu2023texture}, which learns a separate model for each object (e.g., chair, table); MD-ProjTex operates without the need for such class-specific training.
All training-based methods rely on 3D training datasets. Since 3D training data is hard to acquire, they have to use synthetic data, which diverges from the goals of our paper.

\textbf{Novel-view Synthesis.} 
There has been significant progress in novel-view synthesis, where models, given a single input image, can generate novel views by fine-tuning diffusion models on multi-view datasets (e.g., Zero123++ \cite{shi2023zero123++}). While these methods output novel views, the diffusion models lack a mechanism, such as a renderer, to enforce 3D consistency. Recent research, including works like Free3D \cite{zheng2024free3d}, EpiDiff \cite{huang2024epidiff}, Spad \cite{kant2024spad}, IM-3D \cite{melas20243d}, and Carve3D \cite{xie2024carve3d}, has been improving multi-view consistency through attention mechanisms and various optimizations, though it remains an ongoing area of study. While novel-view synthesis excels in single-image reconstruction, our task focuses on generating rich, detailed textures for 3D models, offering the advantage of creating more realistic and immersive virtual environments.

\begin{figure*}[t]
\centering
\includegraphics[width=\linewidth]{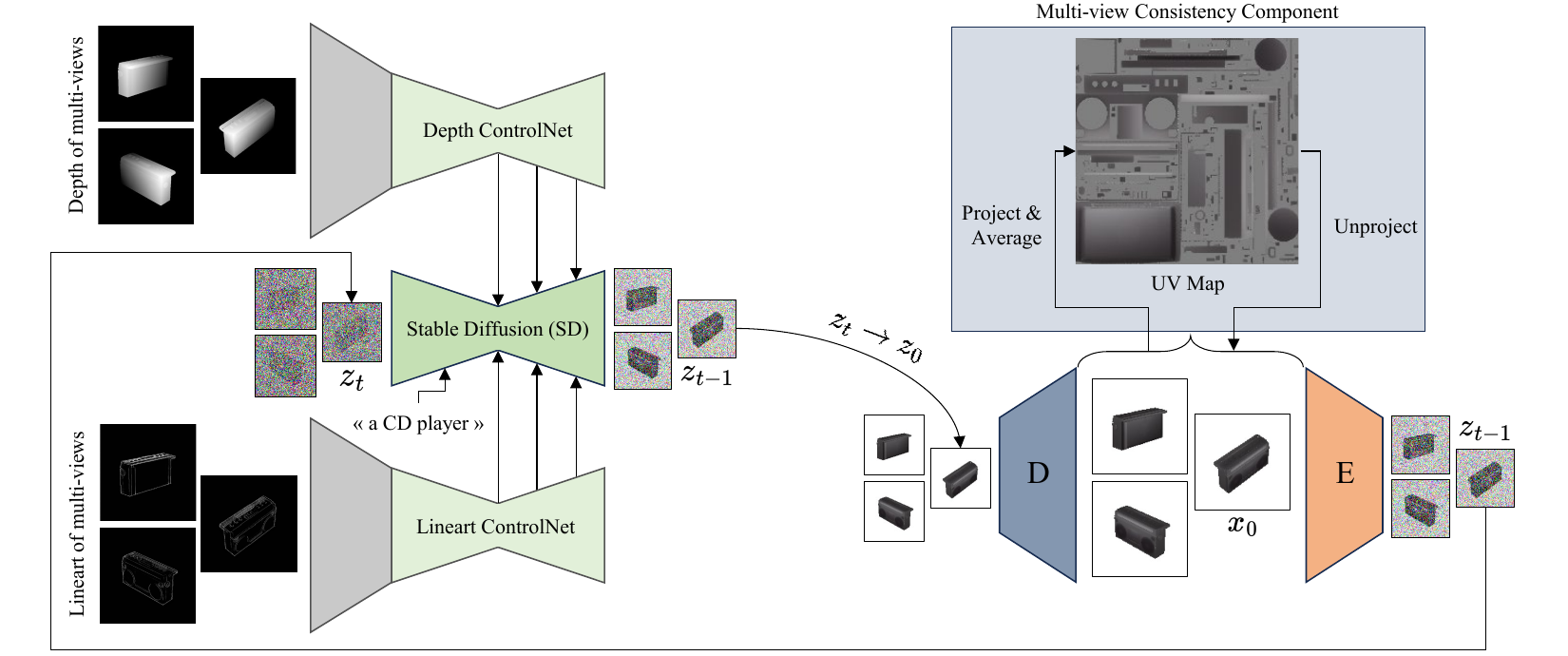}
\caption{We use a latent-based depth and lineart conditioned diffusion model in our pipeline for the denoising steps. At the same time, the multi-diffusion step takes place in the projected textures. To facilitate this process, we incorporate encoder (E), denoising (SD), decoder (D), and projection steps. For simplicity, only three views are shown in the multi-view input visualization. Please note that $z_t$, $z_{t-1}$, and $z_0$ are features in latent space represented as $4\times64\times64$. These features are not directly visualizable. To enhance clarity in the figure, we use downsampled images instead of them. Conversely, $x_0$ exists in image space and is directly visualizable. Starting from $z_T$, we initialize with a normal distribution. Subsequently, the framework employs a pipeline involving denoising, decoding, projection, and encoding for the subsequent steps. 
}
\label{fig:framework}
\end{figure*}

\section{Method}

\subsection{Preliminaries}

We develop our framework on pretrained-diffusion models with a focus on the latent diffusion model \cite{rombach2022high}, chosen for its computational efficiency. 
This model comprises an encoder, denoted as $E$, and a decoder, denoted as $D$.
The encoder is responsible for projecting images into a lower latent dimension, while the decoder reconstructs the images.
The diffusion process, as described in \cite{ho2020denoising}, occurs on the latent codes, denoted as $z_0 = E(x)$, where $x$ represents the input image. At each time step $t$, noise is incrementally introduced to $z_0$ until, after $T$ steps, $z_T$ follows a normal distribution with mean $0$ and identity covariance matrix.

Diffusion models are essentially denoising autoencoders trained to reverse this process, aiming to predict $z_0$ from a denoised version of their input, $z_t$. 

The pre-trained diffusion model we use is depth and lineart conditioned via ControlNet \cite{zhang2023adding}. 
We use 3D geometry and fixed camera viewpoints to generate depth maps that are used to guide the diffusion model.

\subsection{MD-ProjTex Method}

Our aim is to generate outputs that remain consistent across multiple views and can be projected onto UV space to obtain textures. To achieve this, we adopt the multi-diffusion technique \cite{bar2023multidiffusion}, which involves an optimization task that synchronizes multiple diffusion generation processes using a common set of parameters or constraints. This technique has been effectively applied in panorama generation, where multiple generations occur concurrently with shared crops, ensuring coherence across the panorama.
The process takes the average of all its diffusion sample updates for the shared pixels across generations.
Although the multi-diffusion approach has proven effective when applied to latent spaces with shared neighboring crops across generations, our task necessitates its adaptation to the UV space.

\newcommand{\interpfigtt}[1]{\includegraphics[trim=0 0 0cm 0, clip, width=5.5cm]{#1}}
\begin{figure*}[t]
\centering
\addtolength{\tabcolsep}{-5pt}   
\begin{tabular}{ccccccccc}
\\

(a) Iterative Denoising  & (b) Iterative Denoising,   &  (c) Iterative Modified Denoising,  \\
then Decoding.  & Decoding, and Encoding. &   Decoding, and Encoding.\\
\interpfigtt{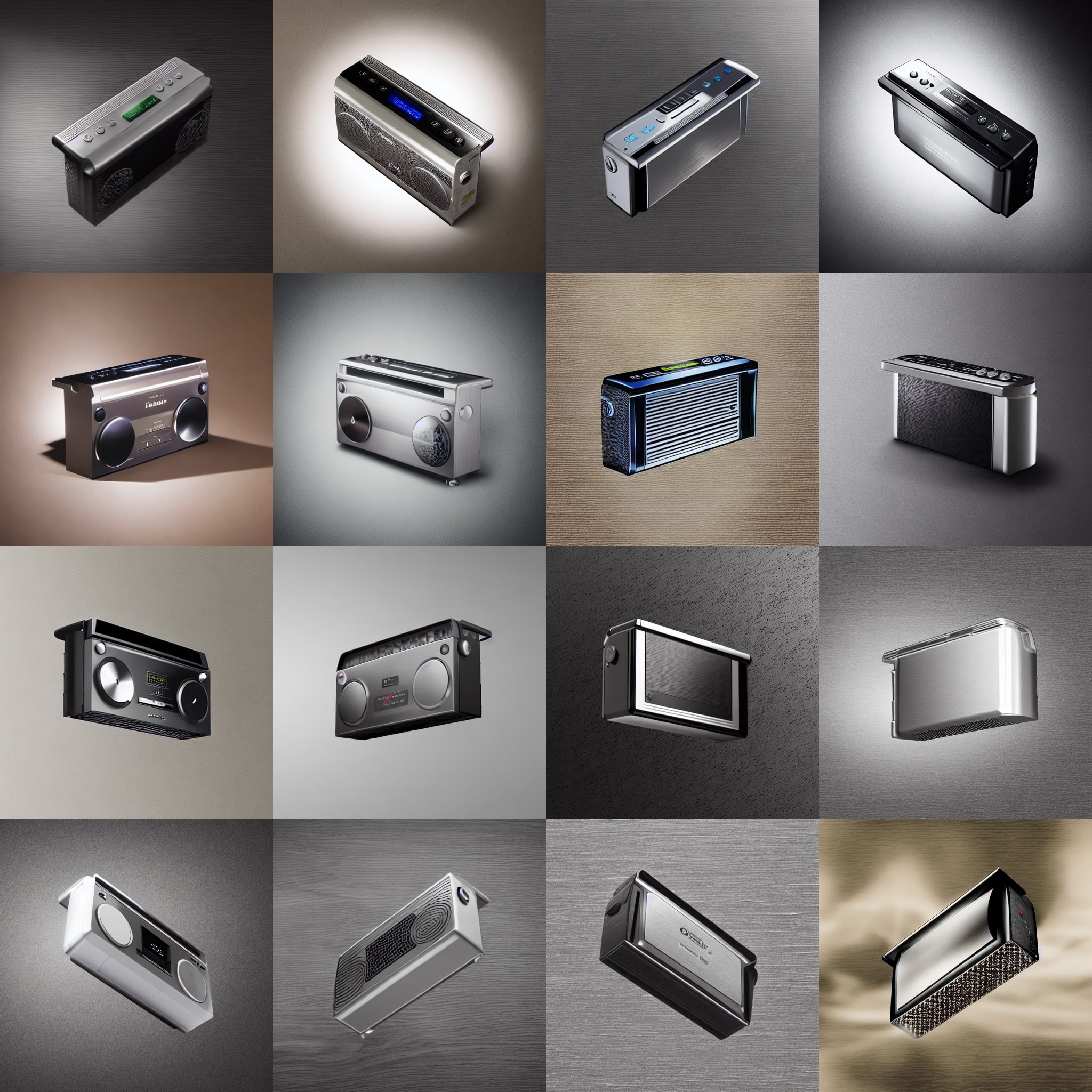} \hspace{1em} &
\interpfigtt{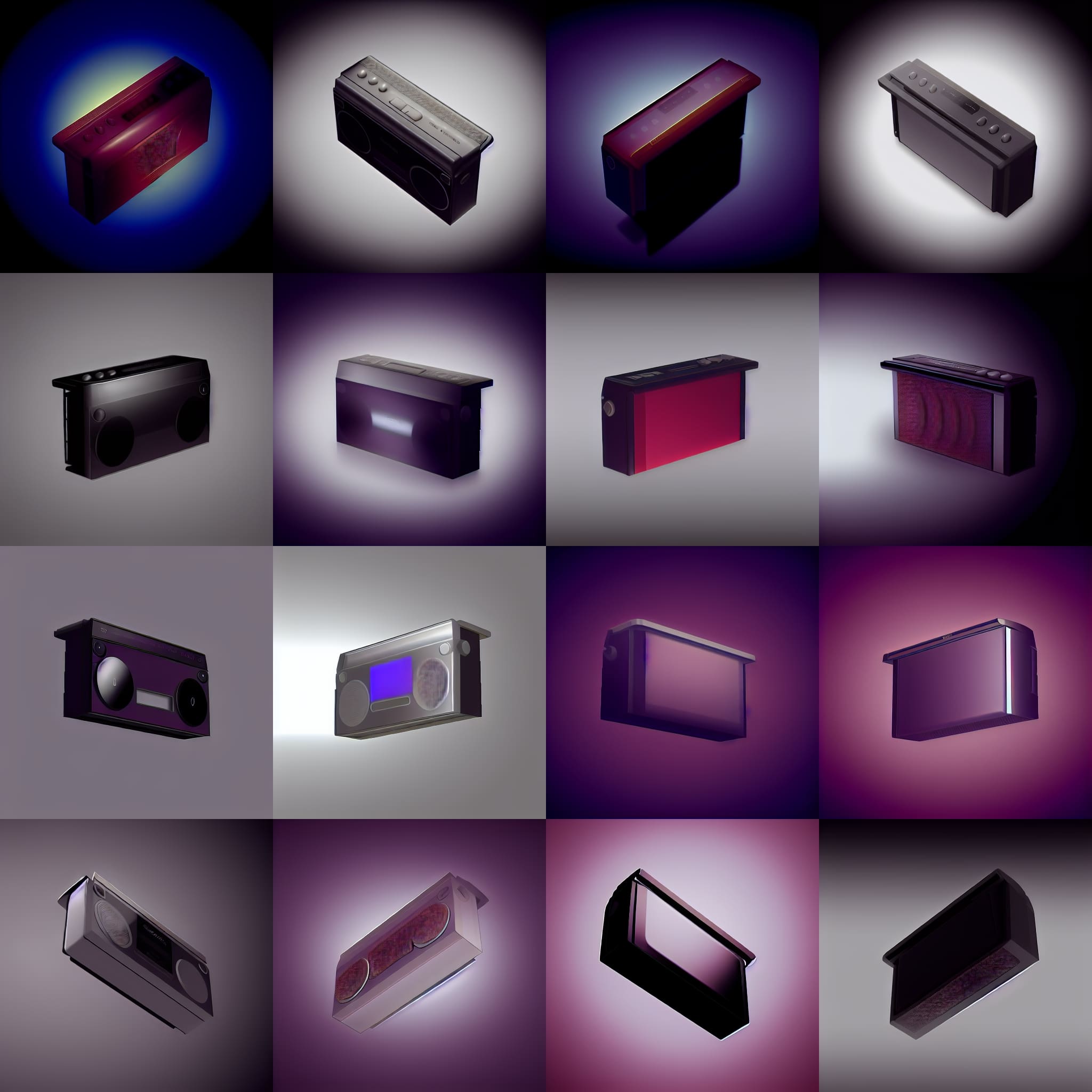} \hspace{1em} &
\interpfigtt{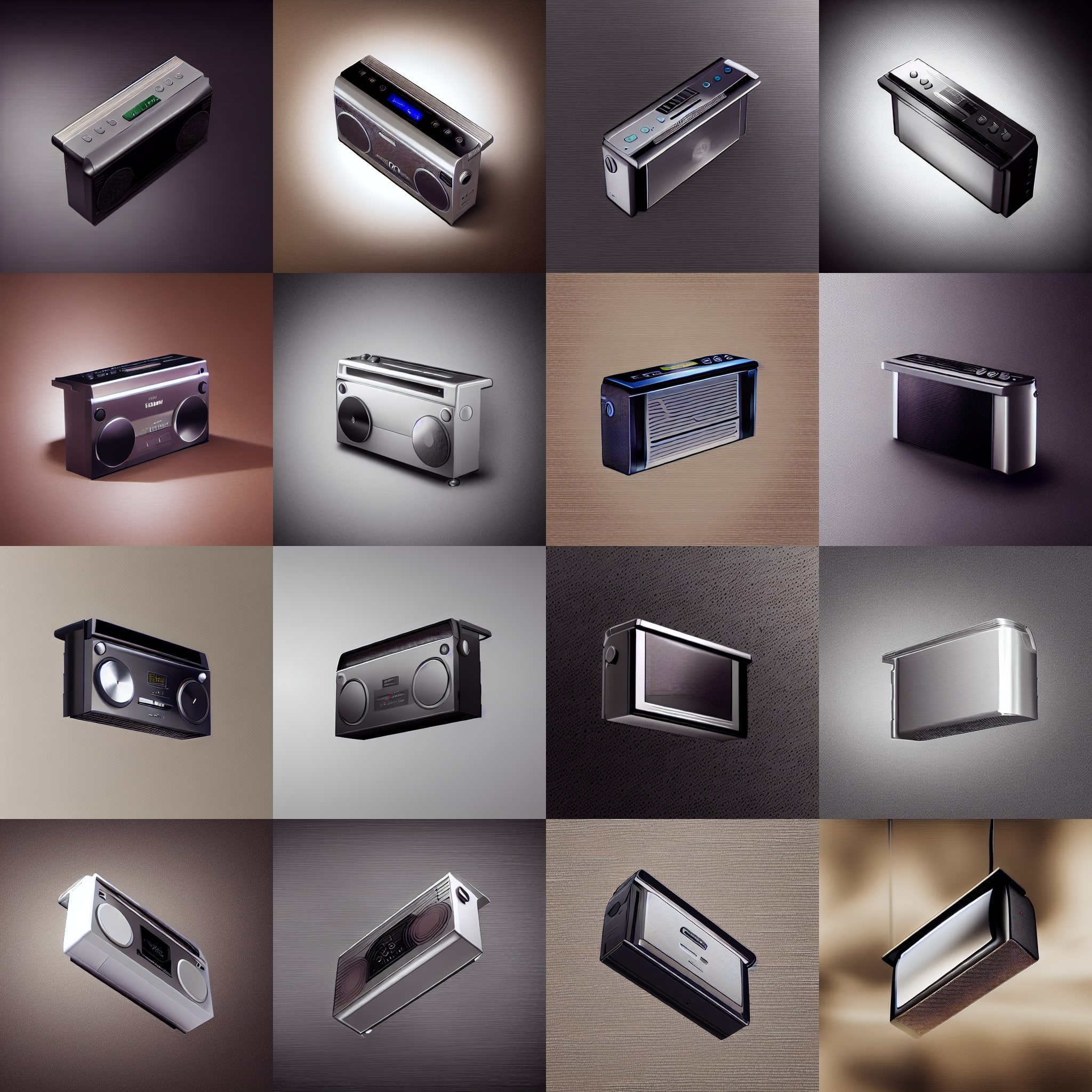}
\\
\end{tabular}
\caption{Ablation results on Stable Diffusion (SD) without the multi-view consistency component are presented in this figure. These experiments aim to demonstrate how a ControlNet-based stable diffusion model generates images. Since the multi-view consistency component is not used in these experiments, we do not expect consistency across views.
(a)  Visual results show the outputs of the original implementation of the SD model, which produces diverse results with a realistic color palette.
(b) In the conventional latent diffusion model setup, denoising steps are applied sequentially in the latent space. Once the latent features are denoised, they are decoded to generate an image. However, in our scenario, during the denoising step, we need to decode the image to apply the multi-view consistency component to the projected textures and then re-encode it to resume the denoising process.
In this experiment, without the multi-view consistency component, we only add the encoder-decoder pipeline. In each denoising step, we decode the prediction, move to the RGB image space, and then encode it back into the latent space. This results in color saturation, leading to purple and pinkish hues. However, for our multi-view consistency component, we are focused on moving to the image space during denoising, as we will use the UV map for averaging.
(c) As a result, we modify the denoising steps as given in Sec.~\ref{sec:subdenosing}. Even with the encoder and decoder in the pipeline, the output images generated resemble those produced by the original implementation.}
\label{fig:method_noising}
\end{figure*}

\subsubsection{Encoder-Decoder-Denoising with Modified Denoising Steps}
\label{sec:subdenosing}
Our first challenge comes from the absence of direct mapping from the encoded latent space to UV space. This mapping is only established after the latents are decoded into image space, as illustrated in Fig. \ref{fig:framework}.
In the conventional setup of latent diffusion models, denoising steps are sequentially applied in the latent space, and once the latent features are denoised via an iterative process, they are decoded to generate an image. However, in our scenario, during the denoising step, we must decode the image to apply multi-diffusion on the projected textures and then encode it back to resume the denoising process.

In each step of this denoising setup, we predict $z_0$ from $z_t$ of all views, decode them to obtain $x_0$,
\begin{equation}
x_0=D(z_0)
\label{eqn:dec}
\end{equation}

Next, the decoded $x_0$'s are projected onto the UV map.
These maps are averaged across views to ensure multi-view consistency, as illustrated in Fig. \ref{fig:framework}.
Afterward, the averaged maps are unprojected back into the image space, continuing the denoising process.
Once unprojected, the images are labeled as $x_0^{md}$. To proceed with further denoising, we then encode these images back into the \( z \)-space, where the denoising process takes place.

 \begin{equation}
z_0'=E(x_0^{md})
\end{equation}

One can then continue the denoising step by the default iterative scheme where latent noise is added to $z'_0$ using:
\begin{equation}
    z_{t-1} = \alpha_{t-1} z'_0 + \sigma_{t-1} \epsilon, \quad \epsilon \sim \mathcal{N}(0, \mathbf{I})
    \label{eqn:denose}
\end{equation}

However, we observe that this framework resulted in color saturation in the generated textures, as will be shown in the ablation study. 
 To better analyze the impact of this behavior, we disabled the multi-view consistency component of the algorithm and performed denoising on single-view images. This experiment is done by predicting $z_0$ from $z_t$, and directly decoding 
 and encoding  them back to obtain $z_0'$ for further denoising by Eq. \ref{eqn:denose}. 
 \begin{equation}
z_0'=E(x_0)
\end{equation}
 
Using this encoder-denoising-decoding scheme, the ControlNet model produces images with saturated colors, as shown in Fig. \ref{fig:method_noising}(b), in contrast to the results from the original implementation, depicted in Fig. \ref{fig:method_noising}(a).
We conclude that this saturation arises due to error accumulation during the encoding and decoding process, leading to trajectory deviations when random latent noise is injected at each step, as described in Eq. \ref{eqn:denose}. A similar phenomenon is observed in experiments on video-to-video translation \cite{yang2023rerender}.

Instead of Eq. \ref{eqn:denose}, we use a guided noise to fix this saturation which we refer to as \textit{Modified Denoising}.
Note that we employ UniPC \cite{zhao2023unipc} scheduler in our implementation, where noise is added as given in Eq. \ref{eqn:denose} and $z_0$ predictions are obtained as outlined in Eq. \ref{eqn:diff_z0}.

\begin{equation}
z_0 = \frac{z_t - \sigma_t\,\epsilon_\theta(z_t, t)}{\alpha_t}
\label{eqn:diff_z0}
\end{equation}

In this equation, the term \(\epsilon_\theta(z_t, t)\) is the noise predicted by the neural network that indicates how much random perturbation is present in \(z_t\), $\sigma_t$ represents the level of noise added, and $\alpha_t$ is defined as $\sqrt{1-\sigma_t^2}$.

In the Modified Denoising method, we reverse this step to estimate the noise direction $\epsilon'$ between $z_t$ and the encoded latent $z_{0}'$:
\begin{equation}
    \epsilon' = \frac{z_t - \alpha_t z_{0}'}{\sigma_t}
\end{equation}

which is then used to take a scheduler step by adding this noise direction between $z_t$ and $z_{0}'$ to obtain $z_{t-1}$:
\begin{equation}
    z_{t-1} = \alpha_{t-1} z_{0}' + \sigma_{t-1} \epsilon' 
\end{equation}

The core idea of this approach is to replace the random latent noise in Eq. \ref{eqn:denose} with a guided noise that points in the same direction as the original noise between $z_t$ and $z'_0$. After obtaining $z_{0}'$ from $z_0$, we return to $z_{t-1}$ by adding this reverse noise direction, helping the process stay close to the original trajectory and reducing deviations. The resulting outputs of this method show that the reconstruction quality is improved significantly by stabilizing the trajectory deviations, as illustrated in Fig.~\ref{fig:method_noising}(c).

\newcommand{\interpfigtn}[1]{\includegraphics[trim=0 0 0cm 0, clip, width=5.5cm]{#1}}
\begin{figure*}[t]
\centering
\addtolength{\tabcolsep}{-5pt}   
\begin{tabular}{ccc}
\\

(a) Normal Maps (colored) & (b) Weight Maps (scale) & (c) Our result \\
\interpfigtn{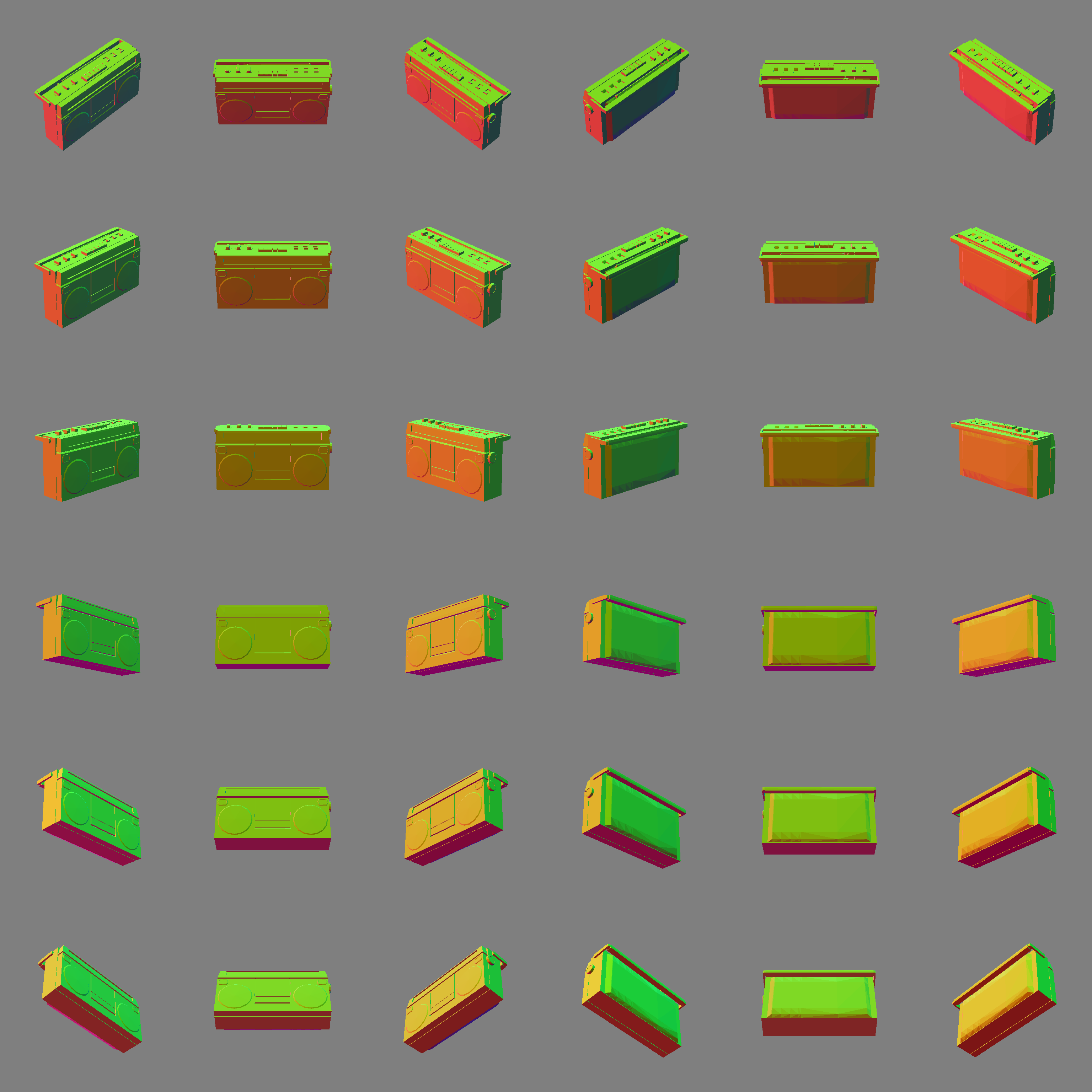} \hspace{1em} &
\interpfigtn{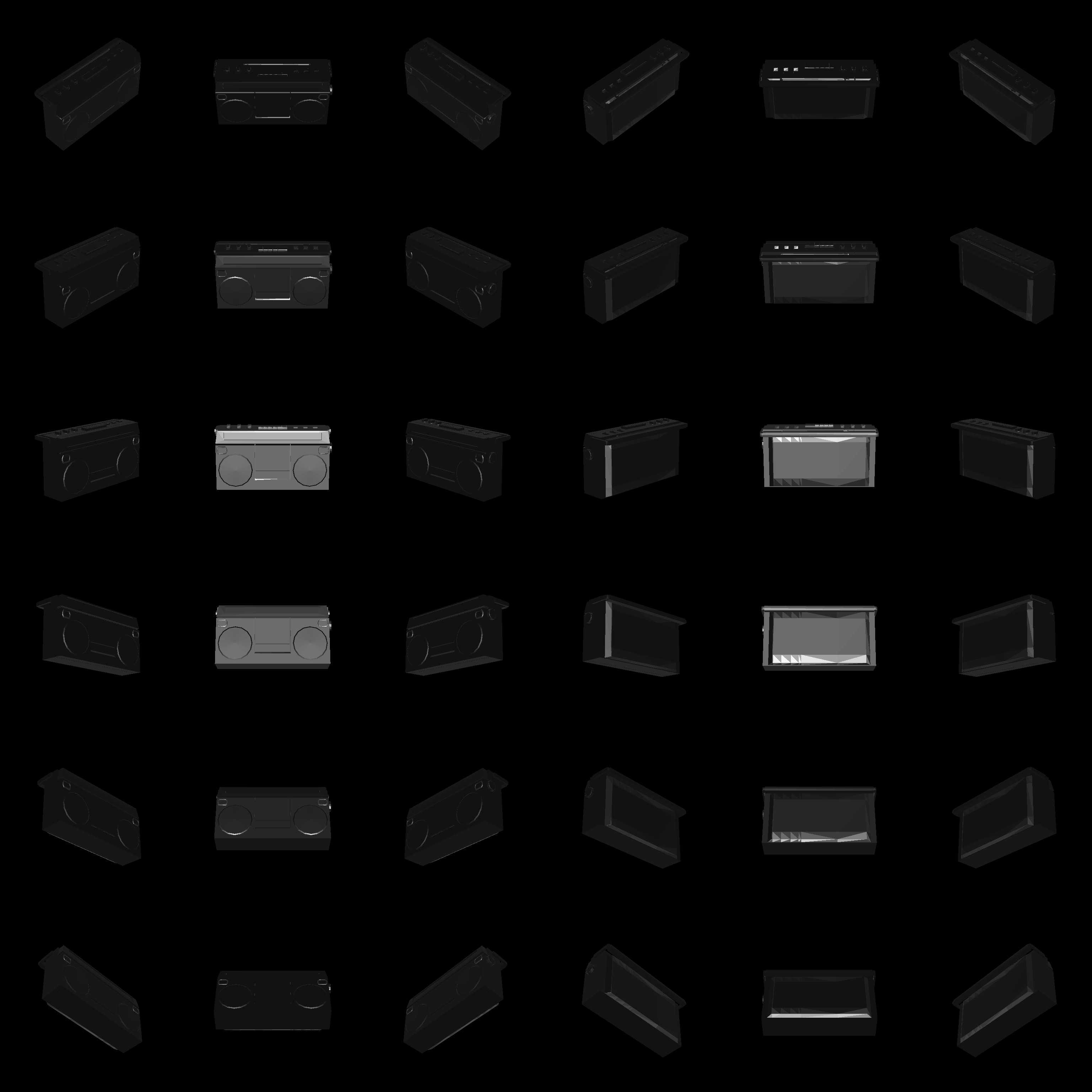} \hspace{1em} &
\interpfigtn{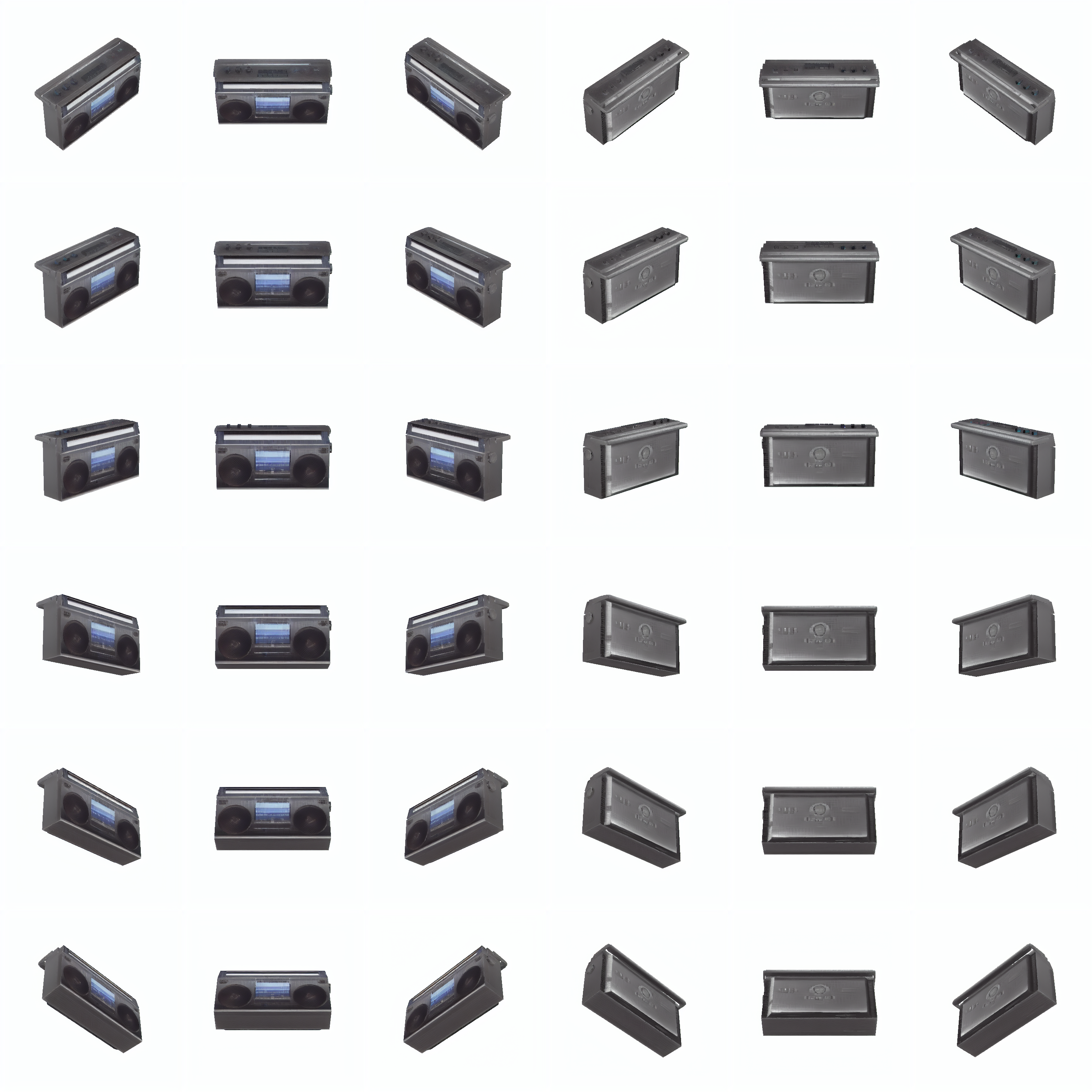} \hspace{1em}
\end{tabular}

\caption{ (a) Visualization of normal maps generated from renderings taken from different camera views. (b) Based on the normal maps, we assign reliability values to various views for a specific pixel on the UV map, with direct views receiving higher reliability scores. (c) Visualization of the texturing results from different camera positions. }
\label{fig:method_normals}
\end{figure*}

\subsubsection{Multi-scale Texture Generation}
The next challenge we observe lies in executing a reliable multi-diffusion step in the texture space. 
One challenge arises from the sparse mapping from image space to textures. In some cases, two views may have pixels that are projected to neighboring pixels but are not shared. Although these pixels are expected to be correlated due to their proximity, they do not appear in the multi-diffusion steps because they are projected to different neighbors.
We propose a multi-scale texture generation approach to address the challenge of sparsity in the mapping from image space to textures, while also introducing normal guidance in the multi-diffusion weighting process to overcome the reliability issue of different views in texture generation.

To implement multi-scale texture generation, we begin by mapping images to lower-resolution textures for earlier time steps. Initially, during the denoising process, we utilize UV maps of size $128\times128$, gradually increasing them to dimensions of $512\times512$. Within our algorithm, at each time step, we project the images onto UV maps, allowing us the flexibility to specify different dimensions for each step.
This way, at the early time steps, where the coarse texture is generated, more pixel sharing is allowed in the UV map.
While this sharing aids in consistency, it is not conducive to achieving finer details.
To achieve fine details in our texture generations, the resolution is increased at later steps. 
These are also the time steps that are known to provide realistic image details. 

\subsubsection{Normal Guidance}
Another issue we encounter here is the inconsistent reliability of each view for any given texture pixel in the UV map.
To weight the multi-diffusion step based on the reliability of the view for a given pixel in the texture space, we utilize the normal maps.
We calculate the difference between the camera space normal and view direction as shown in Fig. \ref{fig:method_normals}(a) for colored and \ref{fig:method_normals}(b) for weight scale visualization. 
We utilize the output of this step to determine the importance of each pixel's contribution to the multi-diffusion process. For instance, in the case of the music player's front view, images in the second column are more reliable as they offer direct viewpoints of the front, particularly evident in the third and fourth samples. Conversely, for the buttons on the top, the first row's second and fifth images provide the clearest perspective, even if they may be visible from multiple viewpoints. It is more probable that the stable diffusion model generates better pixels for these parts when directly viewing them.
With this rationale in mind, we leverage the output of this step to assign weights to each pixel's contribution to the multi-diffusion process. Specifically, for pixels corresponding to the same texture location from different views, we compute their weighted average based on their normal values after applying a softmax operation. To determine the weights, we first compute the angle $\theta_i$ between the surface normal $\mathbf{n}_i$ and the view direction vector $\mathbf{v}_i$ for each contributing pixel $i$. We then define an unnormalized score $s_i = \cos(\theta_i) = \mathbf{n}_i \cdot \mathbf{v}_i$, which measures how directly the surface is facing the camera. The weights are then obtained via a softmax over these scores, as shown in Equation~\ref{eq:softmax}, where $j$ indexes all contributing pixels:

\begin{equation}
w_i = \frac{e^{s_i}}{\sum_j e^{s_j}}.
\label{eq:softmax}
\end{equation}

This approach allows us to achieve both averaging from multiple views and to assign close-to-zero weights to unreliable contributions.

\subsubsection{Camera-view selection} 
Taking a weighted average of pixels, as explained in the previous section, enhances texture details in the generated output, provided that direct views cover most of the object. However, a predefined camera setup does not always guarantee sufficient coverage due to variations in object shapes and surface normals. To address this, we apply weighted clustering on the object's face normals to optimize the camera placement accordingly. Specifically, we use the K-Means algorithm to cluster face normals, assigning face areas as weights to reflect their importance and coverage on the object’s surface. The number of clusters is set to $16$ if it is smaller than the number of faces; otherwise, the number of faces determines the number of clusters. Once clustering is complete, the centroids are selected as target normals, ensuring that cameras are positioned to face them directly. This method increases the coverage of direct views, leading to more detailed texture generation.

 \subsubsection{Post-processing} Finally, there may be a few leftover pixels in the UV maps that have not received projections from any views. Typically, these consist of isolated points surrounded by valid neighboring pixels. To address this, we employ a straightforward post-processing technique using partial convolution operations \cite{liu2022partial} with $3\times3$ filters weighted at $1/9$. Partial convolution employs masks to ensure that empty pixels do not influence the final results. This approach allows us to smoothly fill the empty pixels by averaging their valid neighbors' values.

\section{Experiments}

\textbf{Dataset.} Our method is assessed using a subset of textured meshes sourced from the Objaverse dataset \cite{deitke2023objaverse}. 
We use the same subset as Text2Tex method \cite{chen2023text2tex} which 
comprises 410 high-quality textured meshes spanning 225 categories. 
The original textures are utilized for evaluation purposes.

\textbf{Evaluation.} We assess the quality and diversity of the generated textures using Frechet Inception Distance (FID) \cite{heusel2017gans} and Kernel Inception Distance (KID) \cite{binkowski2018demystifying}. 
The image distributions for these metrics are generated from renders of each mesh, incorporating synthesized textures from 36 fixed viewpoints.
The real image distribution consists of renders of the meshes using ground-truth textures from the same 36 fixed viewpoints.
Note that we do not apply the camera selection method for fair comparisons; however, its advantages are demonstrated in the Ablation Study.
We additionally performed a user study involving 30 samples evaluated by 20 users. Employing an A/B test setup, we present users with text prompts and generations rotated $360^\circ$ along the azimuth. These generations are provided in GIF format, enabling users to easily identify any inconsistencies in the texture.

\textbf{Baselines.} We compare our method with state-of-the-art text-based 3D object texturing methods. Among these methods, CLIP-Mesh \cite{mohammad2022clip} optimizes deformation from a sphere and the surface RGBs. We use the meshes from our dataset only to optimize the texture colors. 
TEXTure \cite{richardson2023texture} and Text2Tex \cite{chen2023text2tex} use a depth-based diffusion model, and starting from a single view, they approach texture generation as an inpainting method, which includes rendering an image from different views and inpainting the pixels that correspond to void textures. 
They both include progressive texture generation and refinement steps and need to go to the image space at each iteration. 
Paint3D \cite{zeng2024paint3d} introduces a two-step texture generation pipeline. This process begins with coarse texture generation using a depth-aware image diffusion model, followed by a refinement stage that employs a UV inpainting model trained on artist-created texture data. While this approach leverages textures crafted by artists, generating such data can be costly and may lack real-world realism. In contrast, our method is founded on image diffusion models, offering a more efficient and potentially more realistic alternative.
Paint-it \cite{youwang2024paint} optimizes the diffusion model and outputs texture maps via synthesis-through-optimization and so resulting in lengthy computation times. Concurrent to our work,  TexPainter \cite{zhang2024texpainter} proposes running a denoising process from multiple views and blending them into a common color-space texture image. To achieve this, the method optimizes the latent space of the decoder via backpropagation, which also results in long computational times.

\newcommand{\interpfigmm}[1]{\includegraphics[trim=3cm 2.5cm 3cm 2.5cm, clip, width=2.2cm]{#1}}
\begin{figure*}[t]
\scalebox{0.8}{
\centering
\begin{tabular}{cccccccccccccccccc}

\rotatebox{90}{~CLIP-Mesh} &
\interpfigmm{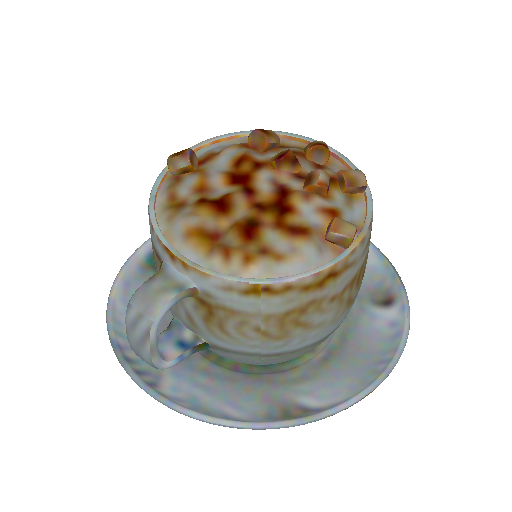} &
\interpfigmm{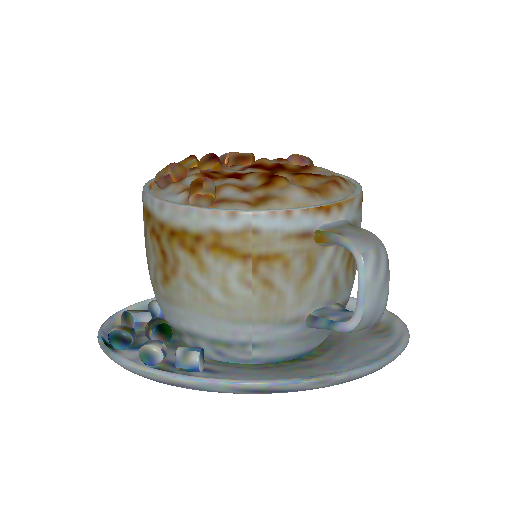} &
\interpfigmm{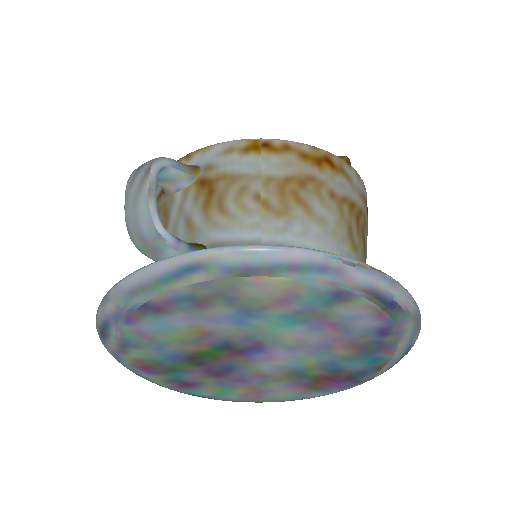} &
\interpfigmm{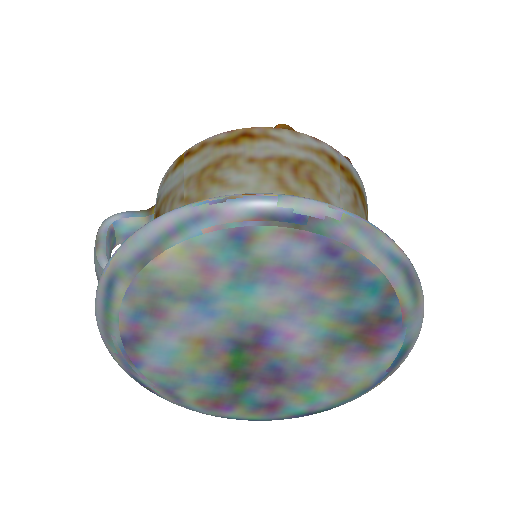} & 
\interpfigmm{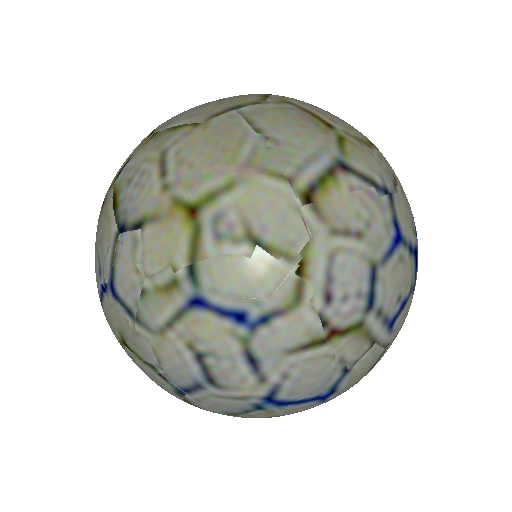} &
\interpfigmm{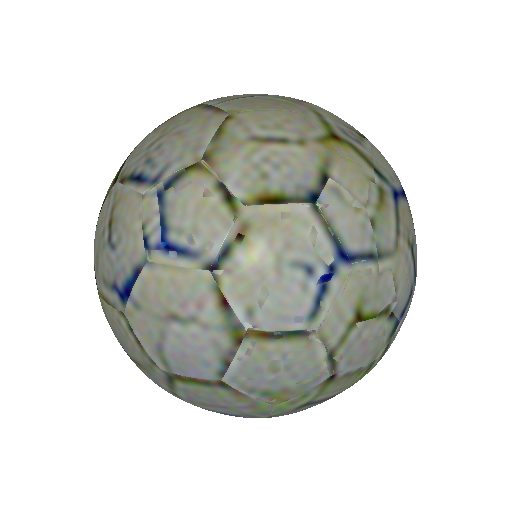} &
\interpfigmm{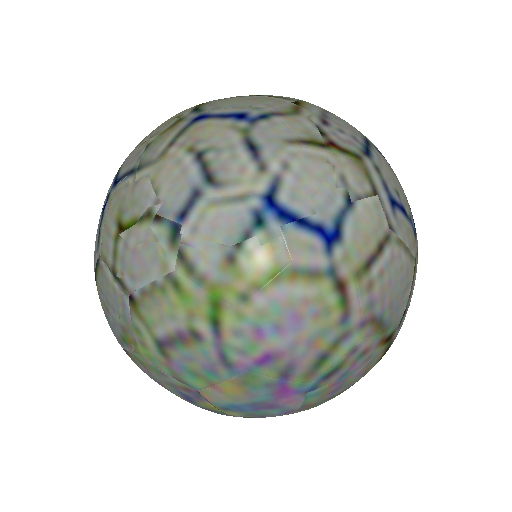} &
\interpfigmm{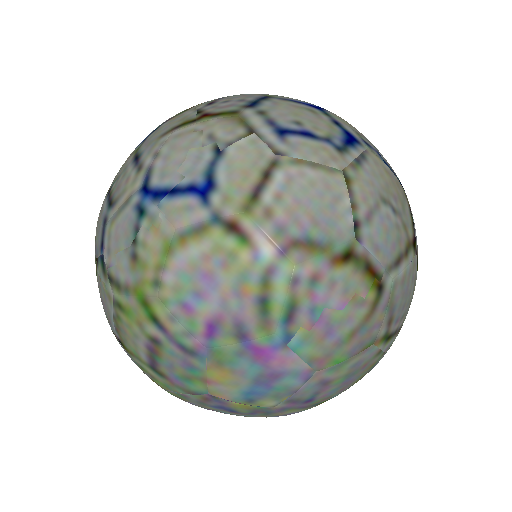}
\\
\rotatebox{90}{~~~TEXTure} &
\interpfigmm{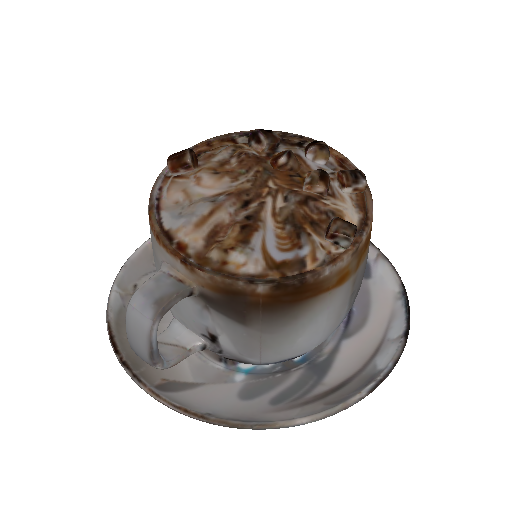} &
\interpfigmm{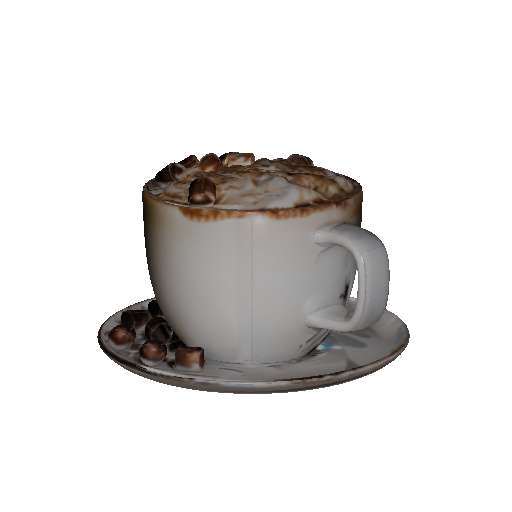} &
\interpfigmm{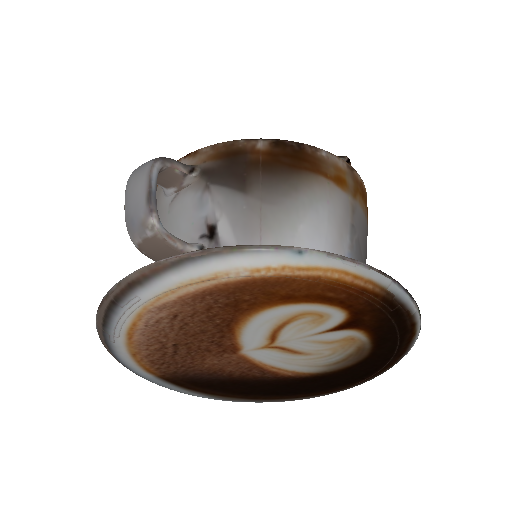} &
\interpfigmm{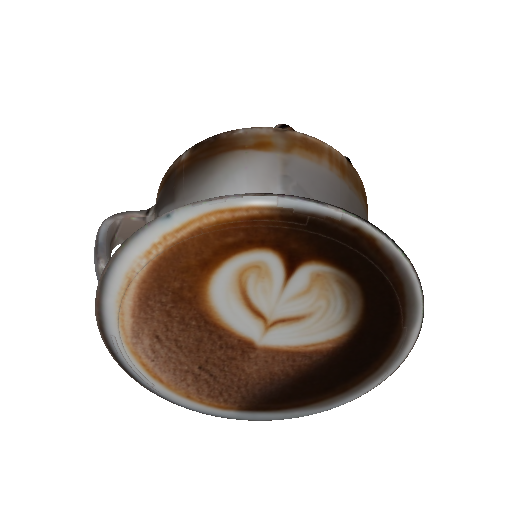} & 
\interpfigmm{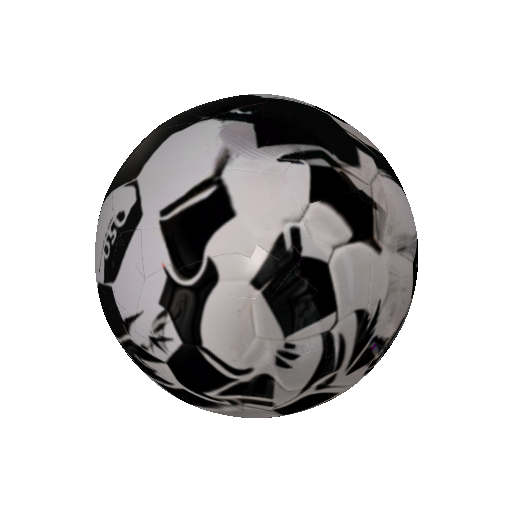} &
\interpfigmm{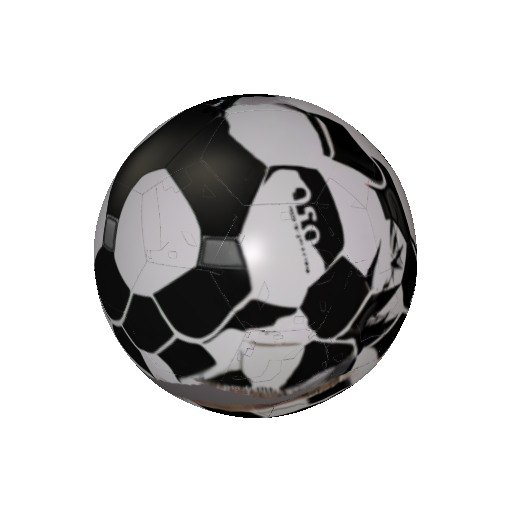} &
\interpfigmm{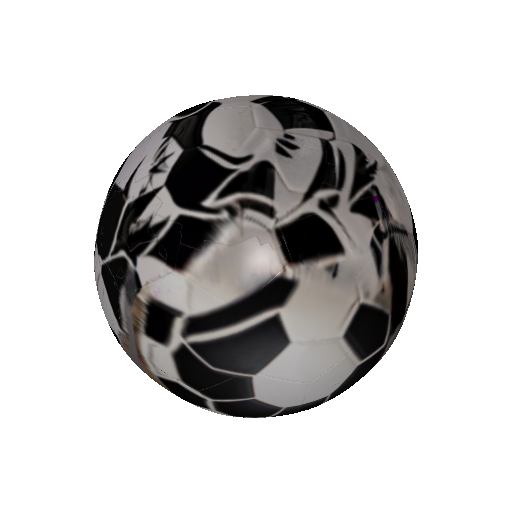} &
\interpfigmm{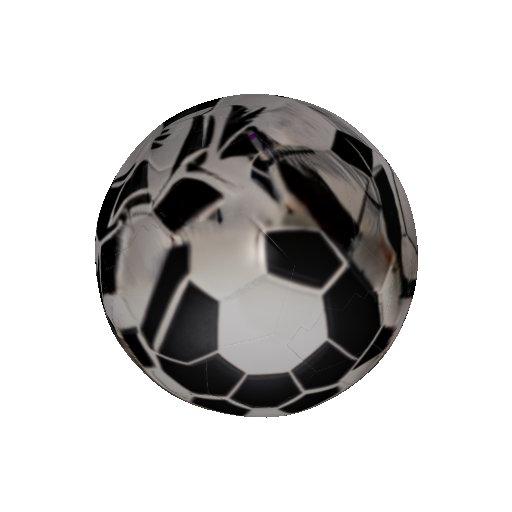} 
\\
\rotatebox{90}{~~~Text2Tex} &
\interpfigmm{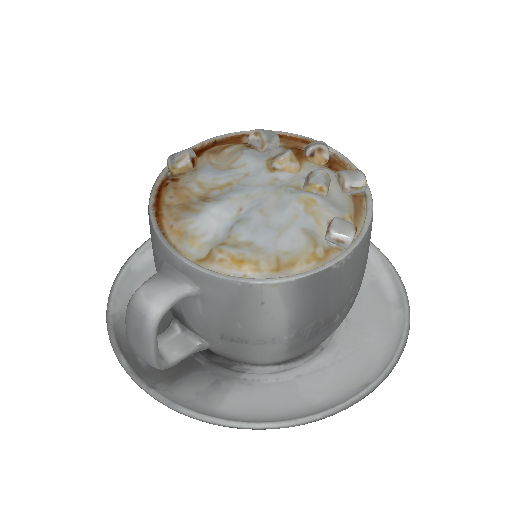} &
\interpfigmm{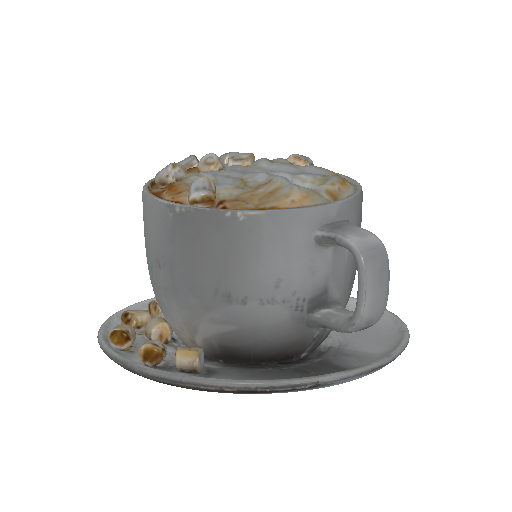} &
\interpfigmm{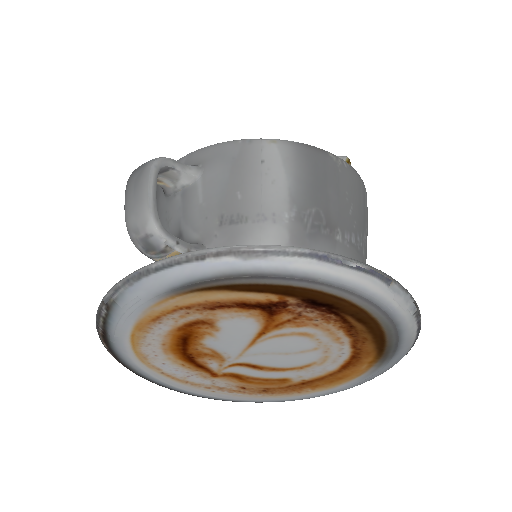} &
\interpfigmm{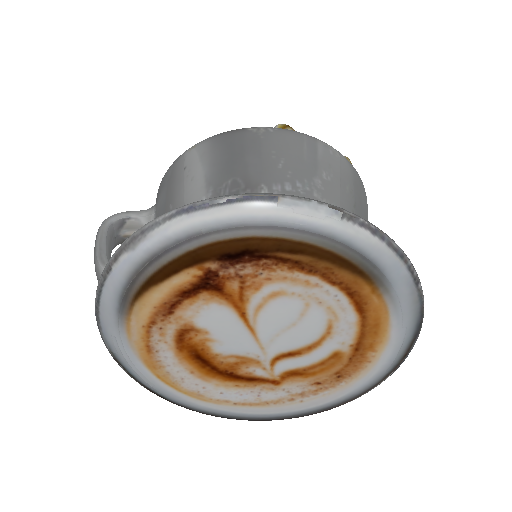} & 
\interpfigmm{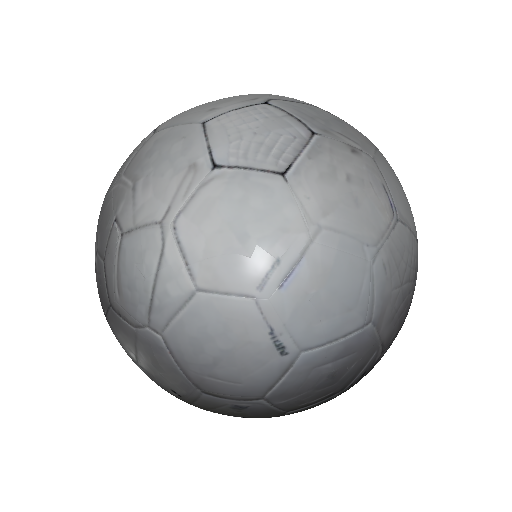} &
\interpfigmm{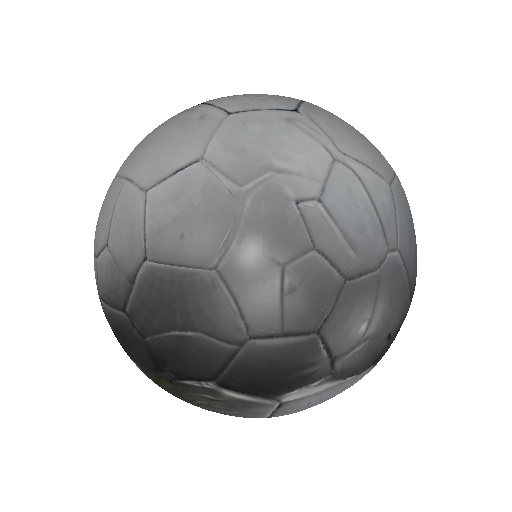} &
\interpfigmm{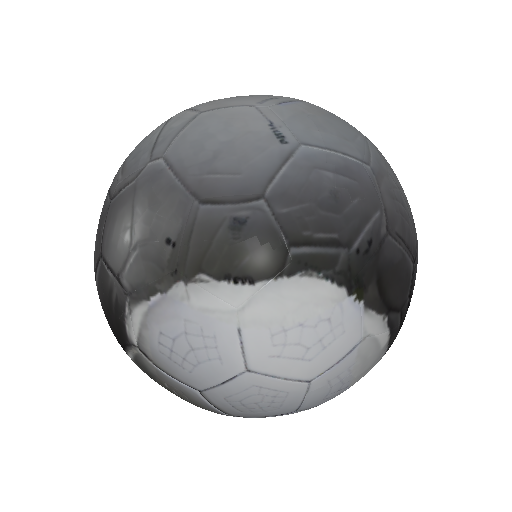} &
\interpfigmm{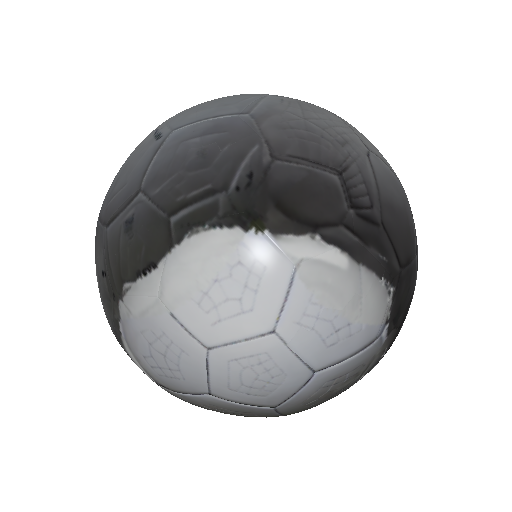}
\\
\rotatebox{90}{~~~~Paint3D} &
\interpfigmm{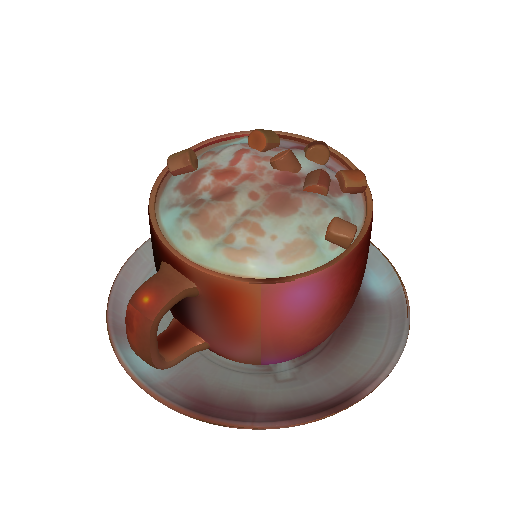} &
\interpfigmm{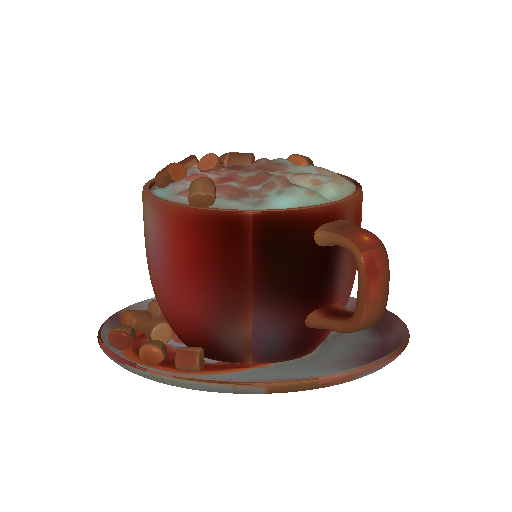} &
\interpfigmm{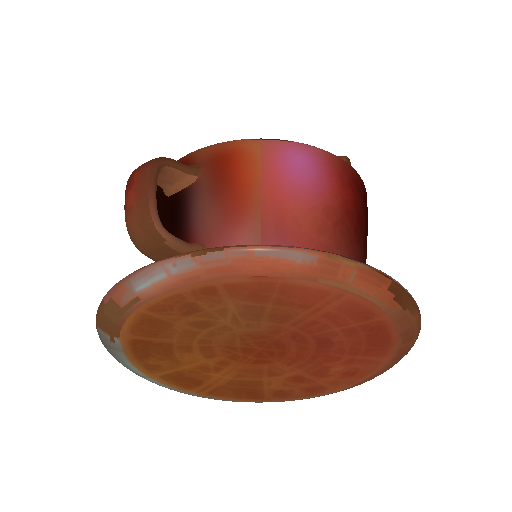} &
\interpfigmm{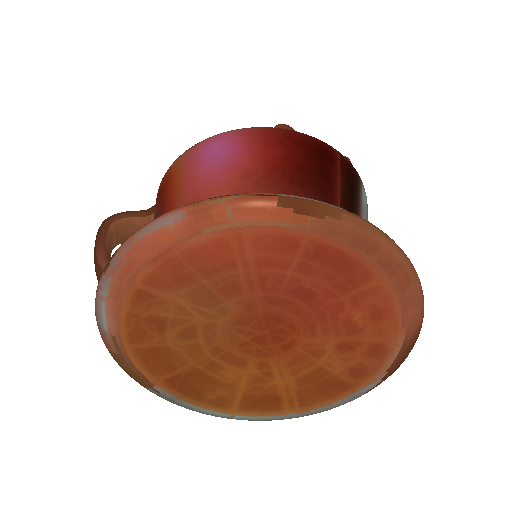} & 
\interpfigmm{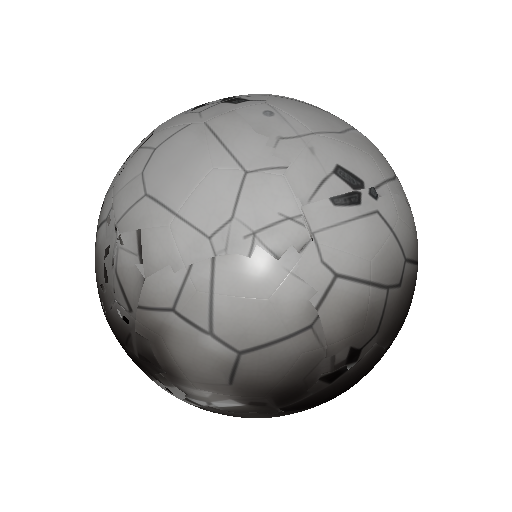} &
\interpfigmm{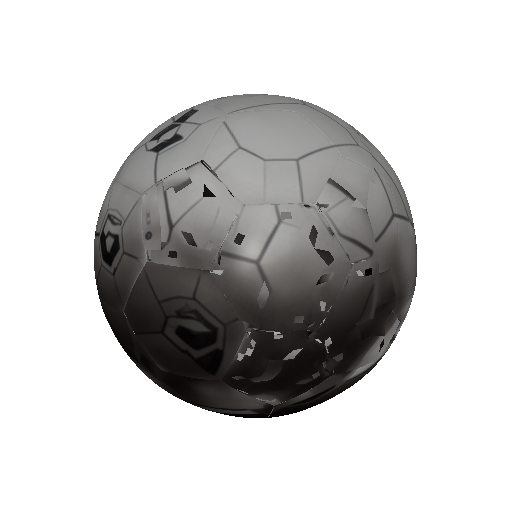} &
\interpfigmm{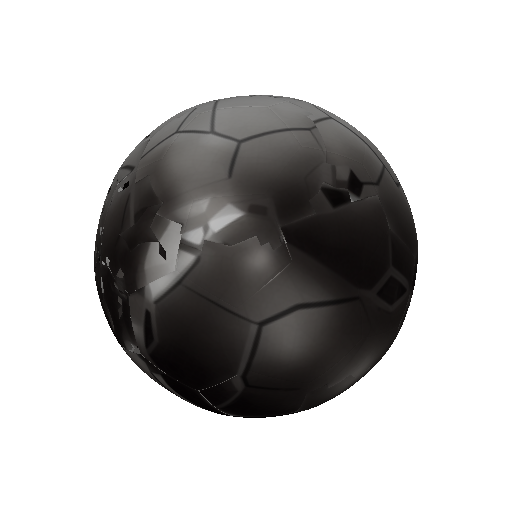} &
\interpfigmm{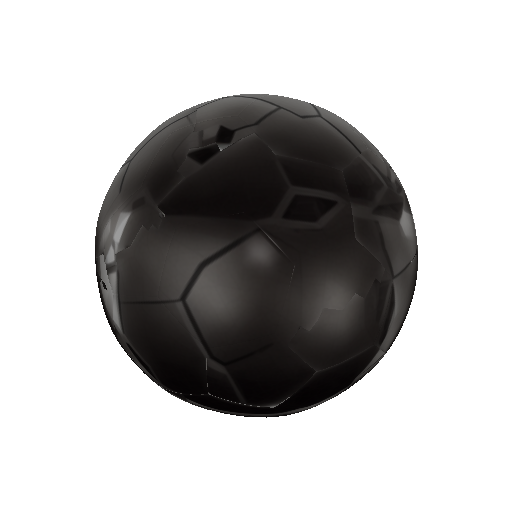}
\\
\rotatebox{90}{~~TexPainter} &
\interpfigmm{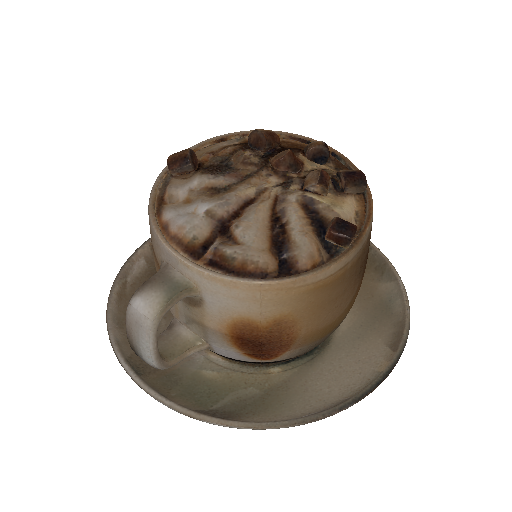} &
\interpfigmm{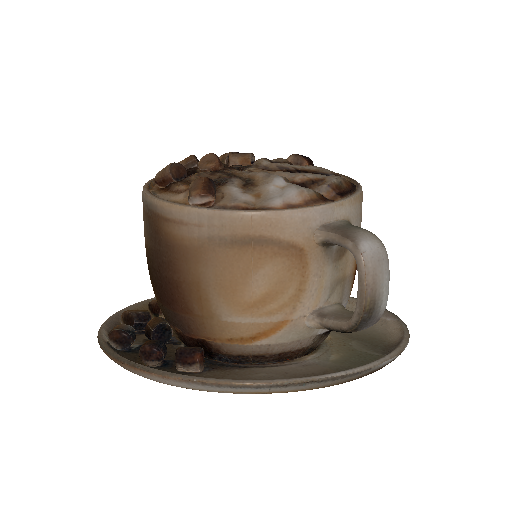} &
\interpfigmm{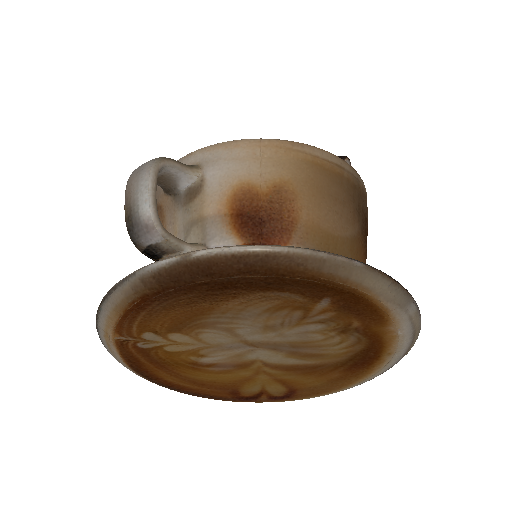} &
\interpfigmm{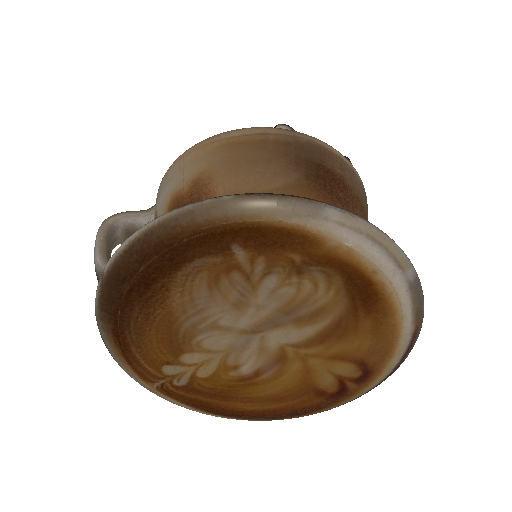} & 
\interpfigmm{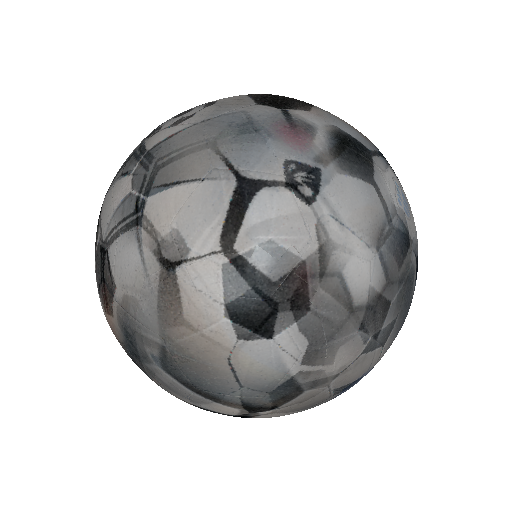} &
\interpfigmm{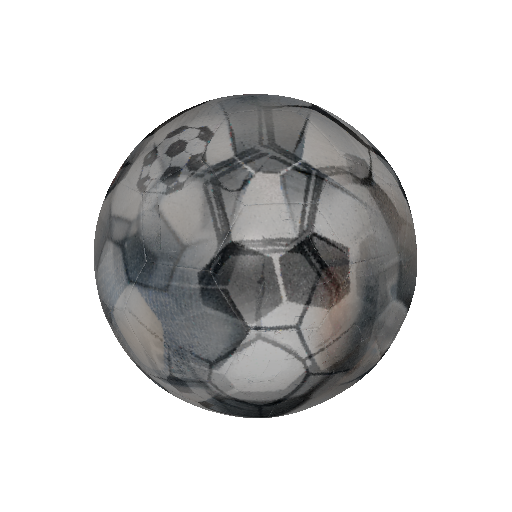} &
\interpfigmm{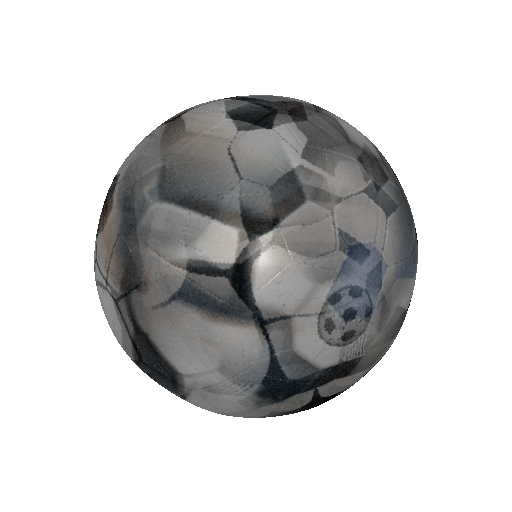} &
\interpfigmm{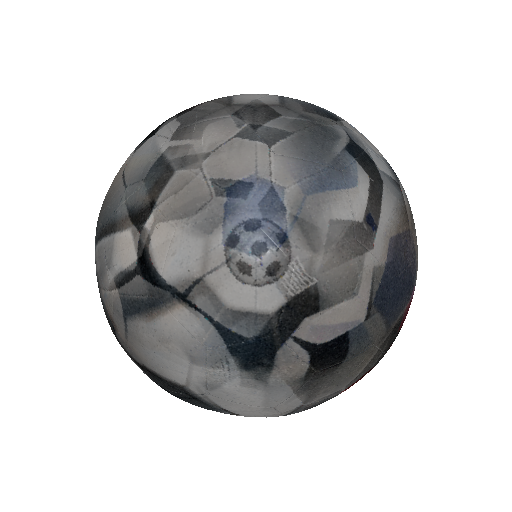}
\\
\rotatebox{90}{~~~~Paint-it} &
\interpfigmm{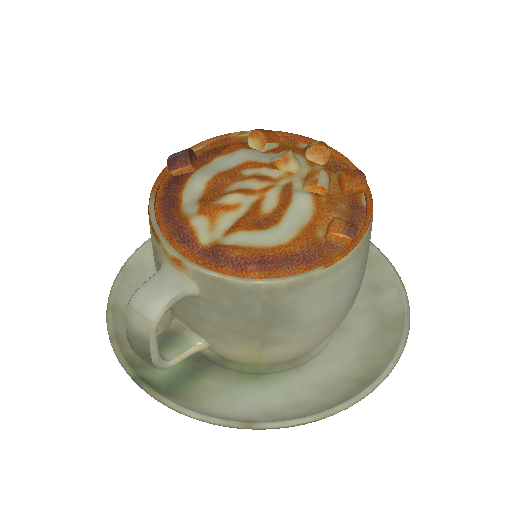} &
\interpfigmm{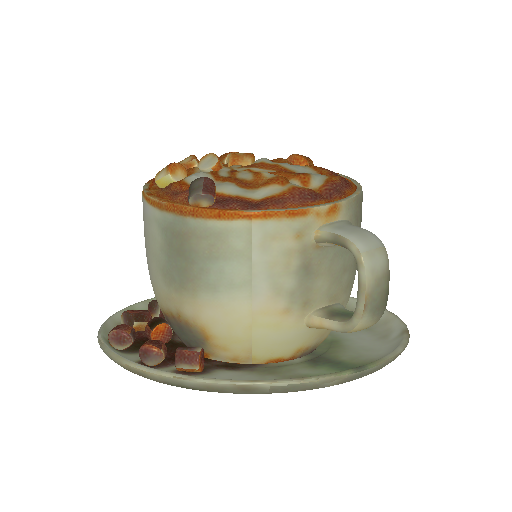} &
\interpfigmm{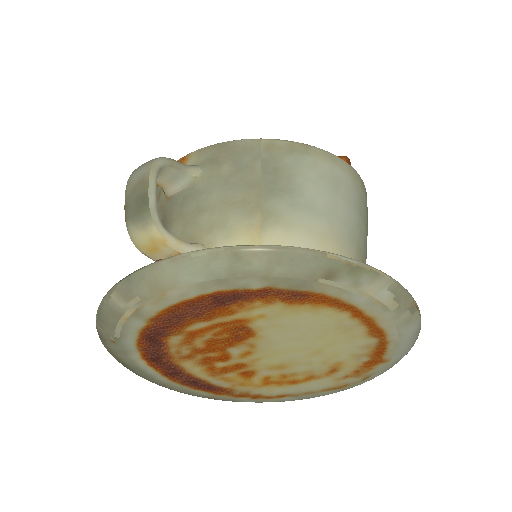} &
\interpfigmm{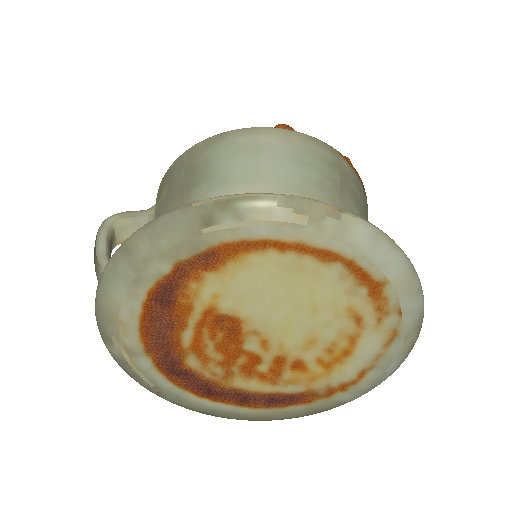} & 
\interpfigmm{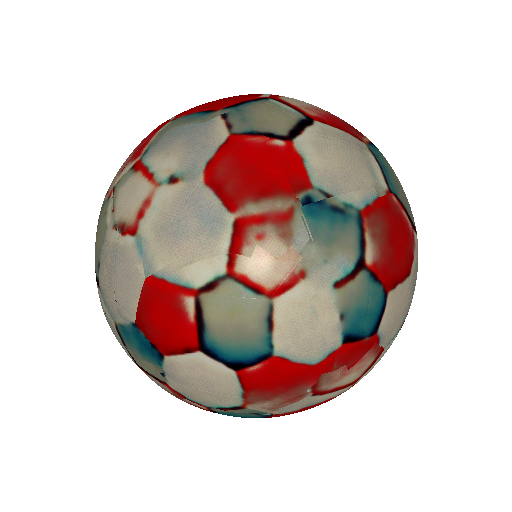} &
\interpfigmm{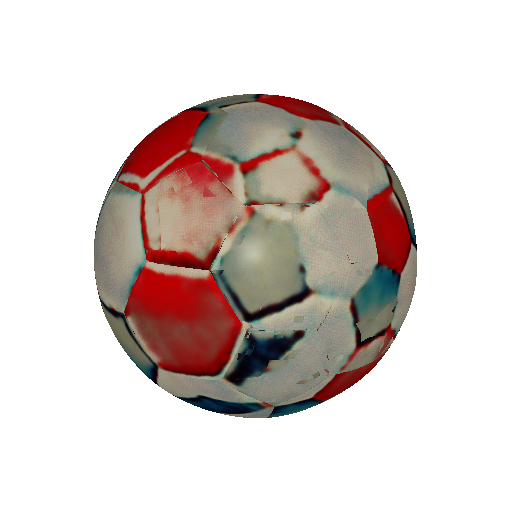} &
\interpfigmm{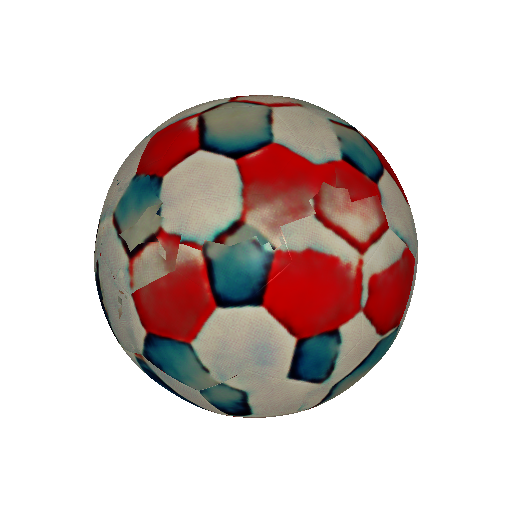} &
\interpfigmm{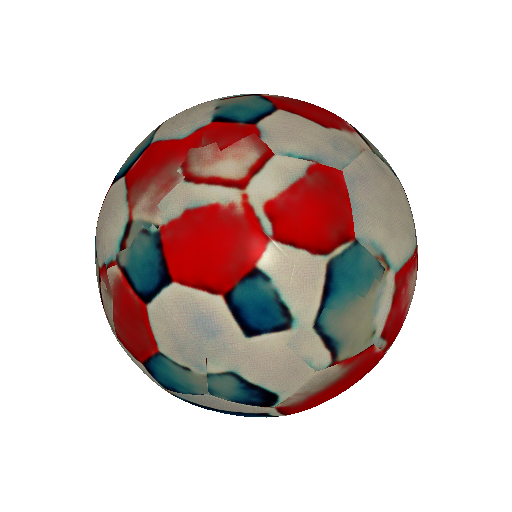}
\\
\rotatebox{90}{~~~~~ Ours} &
\interpfigmm{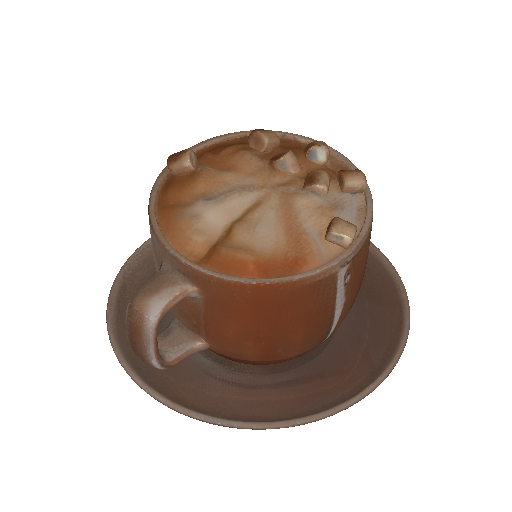} &
\interpfigmm{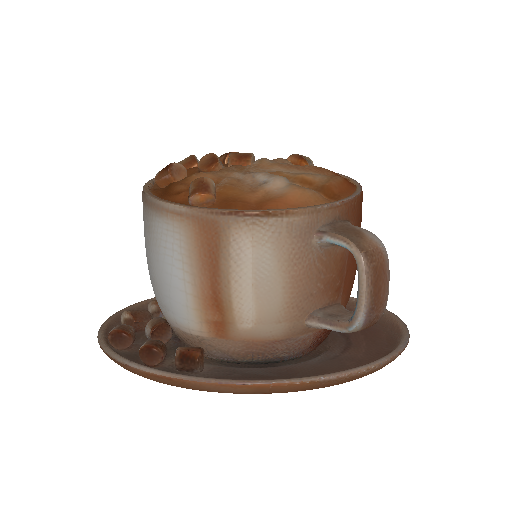} &
\interpfigmm{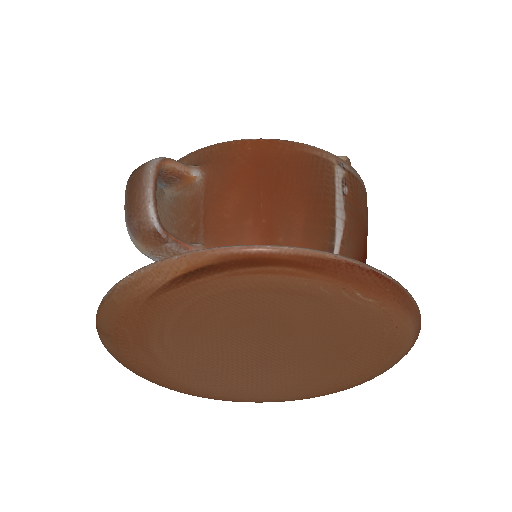} &
\interpfigmm{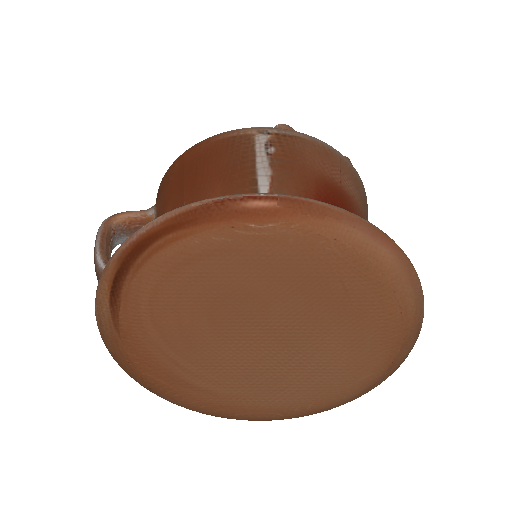} & 
\interpfigmm{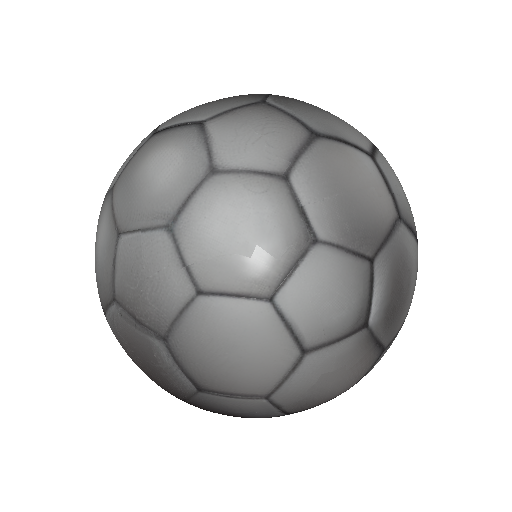} &
\interpfigmm{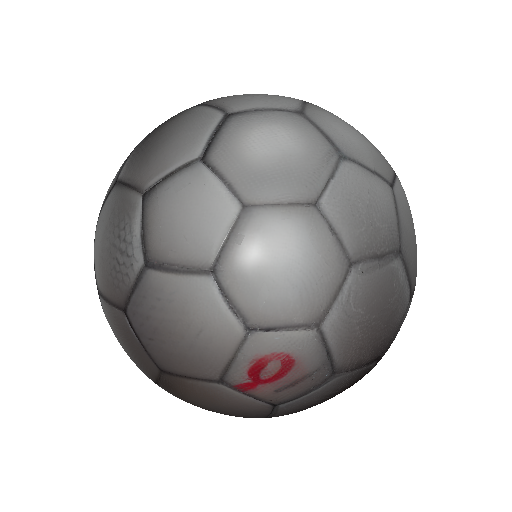} &
\interpfigmm{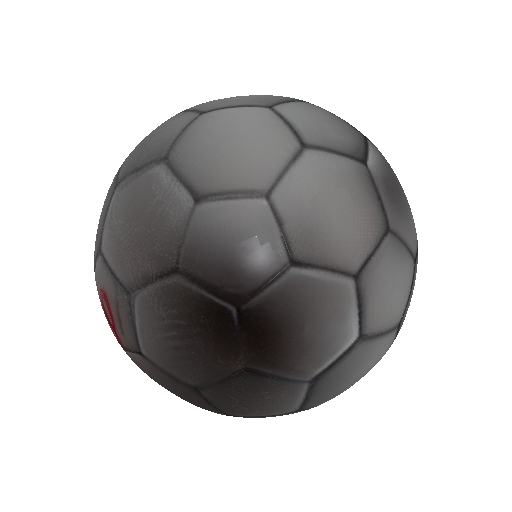} &
\interpfigmm{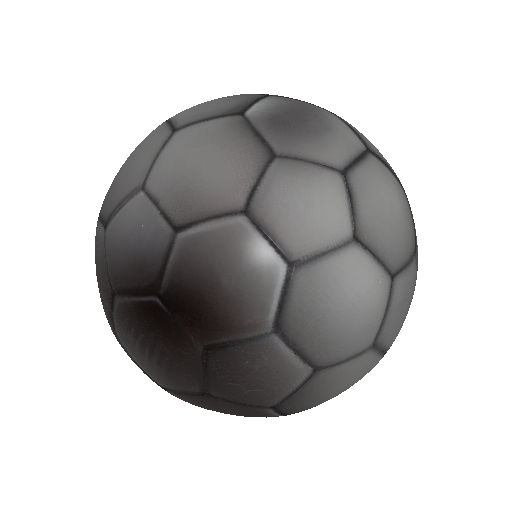} 
\\
& \multicolumn{4}{c}{\textit{A cappuccino}} & \multicolumn{4}{c}{\textit{A soccer ball}}
\\
\end{tabular}
}
\caption{Qualitative results of our and competing methods with multi-view renderings. }
\label{fig:results_simple}
\end{figure*}

\begin{figure*}[h!]
\centering
\includegraphics[width=0.9\linewidth]{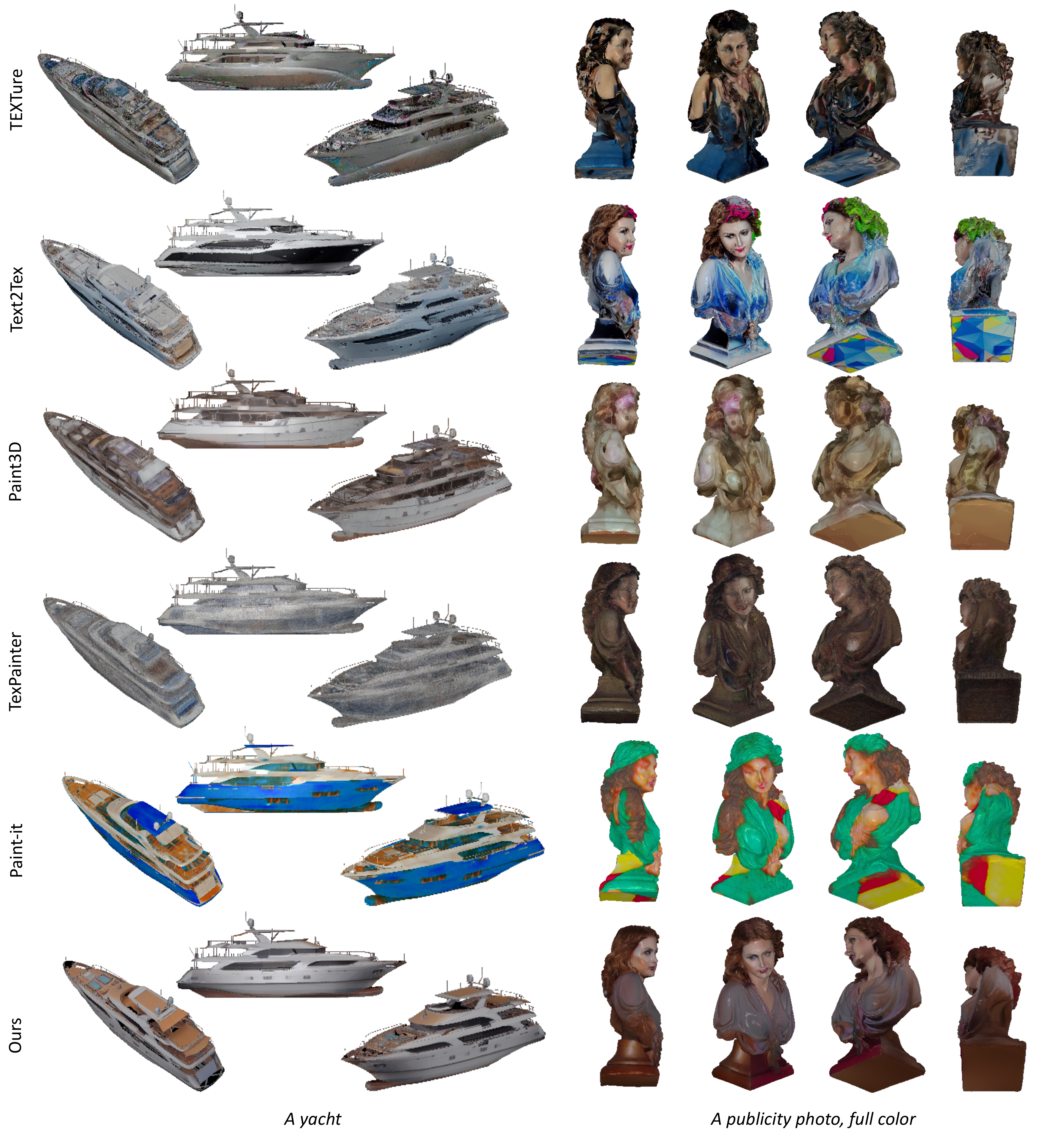}
\caption{Qualitative results of our and competing methods with multi-view renderings.}
\label{fig:results_complex}
\end{figure*}

\begin{table*}[t!]
\centering
\caption{Quantitative results of our and competing methods on the objaverse-benchmark dataset. }
\begin{tabular}{|l|c|c|r|c|}
\hline
Models  & {FID} $\Downarrow$ & KID ($\times 10^{-3}$) $\Downarrow$ & {Runtime (min)} $\Downarrow$ &  Preference of Ours  \\ \hline
CLIP-Mesh \cite{mohammad2022clip} & 36.73 & 7.81 & 41.3 & \textbf{87.4}\\
TEXTure \cite{richardson2023texture} & 25.47 & 4.36 & 6.5 &  \textbf{66.3} \\
Text2Tex \cite{chen2023text2tex} & 22.66 & 3.21 & 30.4 & \textbf{58.4} \\
Paint3D \cite{zeng2024paint3d} & 22.30 & 3.78 & 2.7 & \textbf{62.0} \\
TexPainter \cite{zhang2024texpainter} & 18.92 & 3.69 & 71.5 & \textbf{67.4} \\
Paint-it \cite{youwang2024paint} & 25.92 & 4.02 & 14.4 & \textbf{86.7} \\
\hline
Ours & 16.63 & 1.73 & \textbf{2.6} & - \\ 
\hline
\end{tabular}
\label{table:results_sota_quantitative}
\end{table*}

\textbf{Implementation Details.}
In our setting, for the 3D rendering, we use PyTorch3D \cite{ravi2020accelerating}.
As for our image generator, we utilize a pretrined Weighted Multi-ControlNet model for Stable Diffusion v1.5 that guides the image generation process based on given depth and lineart conditions. The weights of both ControlNet models are set to $0.5$ for balancing the guidance. We generate 36 views simultaneously for each object. There is no tuning for the view selections, the same camera parameters are used in our experiments. The view images generated by the Stable Diffusion model are in $512\times512$ dimension. UV mapping in the multi-view consistency component is applied on $128\times128$, $256\times256$, and $512\times512$ textures. A weighted average is taken at each timestep when obtaining the projected $x_0$ image. Initially, all weights are given to the $128\times128$ texture. Up to the $0.3$ of the timesteps, the weights are linearly transferred to the $256\times256$ texture. In the following timesteps, weights are transferred with a similar method such that the weights of $256\times256$ and $512\times512$ textures have $0.4$ and $0.6$ weights at the end, respectively.

\textbf{Results.}
We provide the quantitative and qualitative results in Table \ref{table:results_sota_quantitative} and Fig. \ref{fig:results_simple}, respectively. 
Following previous evaluation set-ups, we feed template texts
"a \textless category\textgreater" to the models for each object.
In quantitative metrics, our method surpasses the FID and KID scores of competing methods, achieving this with an impressive speed-up compared to others.
Such improvements in FID and KID mainly come from the multi-view consistency of our generations and the realism we achieve compared to the others. 
Our model is faster because we do not have length optimizations or sequential generation of each view.
The run-times are measured on Tesla T4, single GPU.
In our user study, our model was preferred over competing methods significantly.

Our visual comparisons reveal that competing methods often produce noisy patterns and struggle to generate a consistent texture, especially when rendered from multiple views, both on simple objects and complex objects, as shown in Fig. \ref{fig:results_simple} and \ref{fig:results_complex}, respectively. 
One limitation we have observed in other methods is their difficulty in producing reasonable outputs for challenging views, such as the bottom of the coffee plate, as depicted in Fig. \ref{fig:results_simple}. When prompted with the term "cappuccino", these models tend to fill even the back of the plate with a cappuccino.
In both numerical and visual comparisons, across both simple and complex shapes, we find that our method produces better results while running faster.

In Fig. \ref{fig:results_ours_multiprompt}, we demonstrate that our method produces diverse results, generating multi-view consistent outputs that are in line with the given prompts. Our approach can also be combined with adapters designed for image diffusion models. As illustrated in Fig. \ref{fig:results_ip_adapter}, we demonstrate the integration of the IP-Adapter \cite{ye2023ip} with our method. By utilizing the IP-Adapter, we can incorporate style images, allowing us to guide texture generation not only through text conditioning but also via the style image.

\newcommand{\interpfigte}[1]{\includegraphics[trim=5cm 3cm 5cm 3cm, clip, width=1.4cm]{#1}}
\begin{figure*}[t]
\centering
\addtolength{\tabcolsep}{-0.2pt}  
\begin{tabular}{ccccccccccc}
\interpfigte{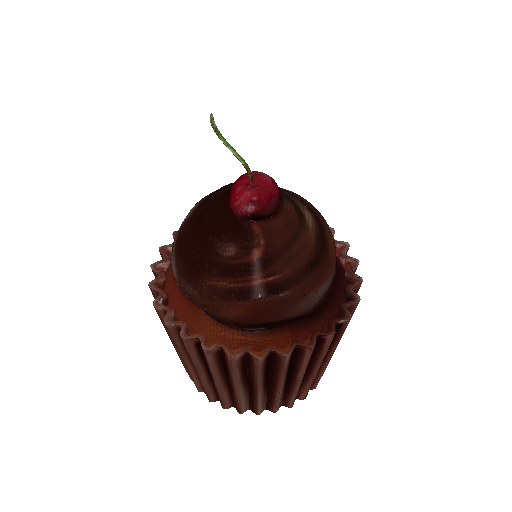} &
\interpfigte{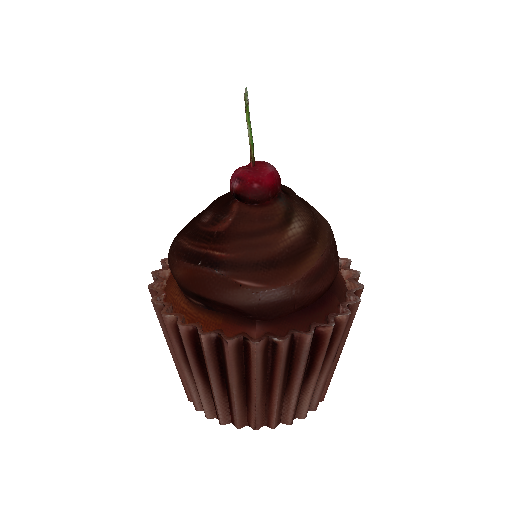} &
\interpfigte{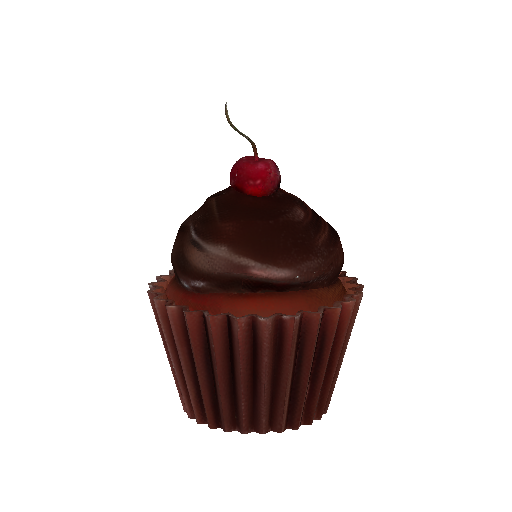} &
\interpfigte{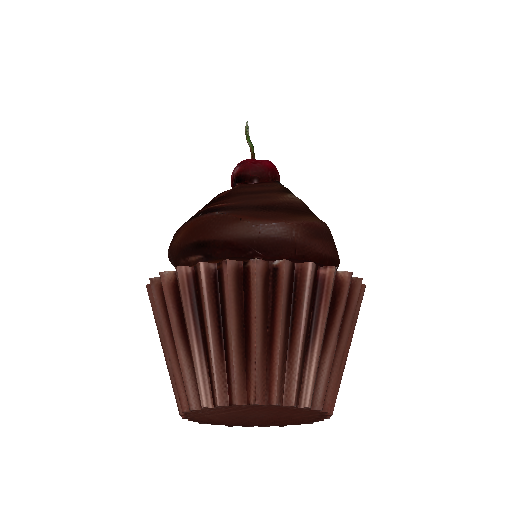} &
\interpfigte{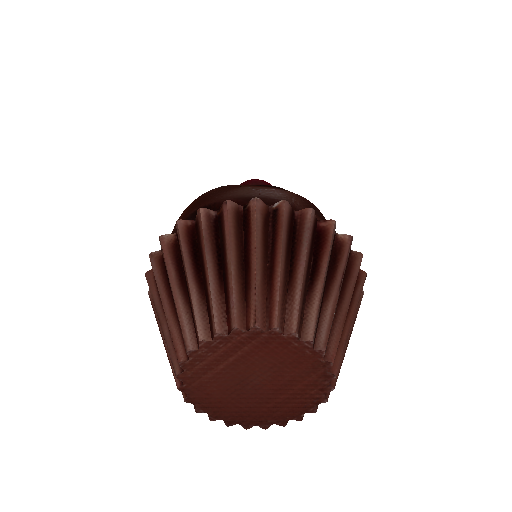} &
\interpfigte{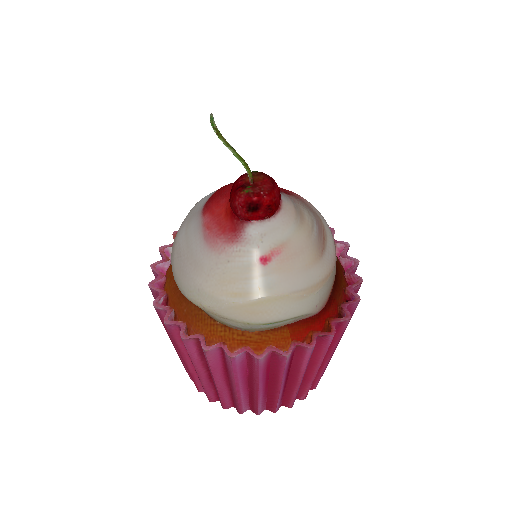} &
\interpfigte{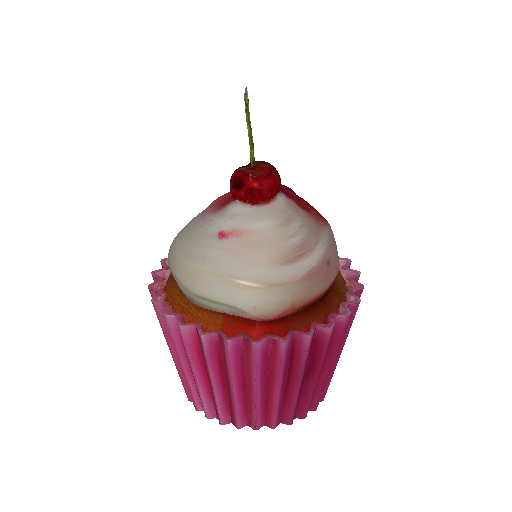} &
\interpfigte{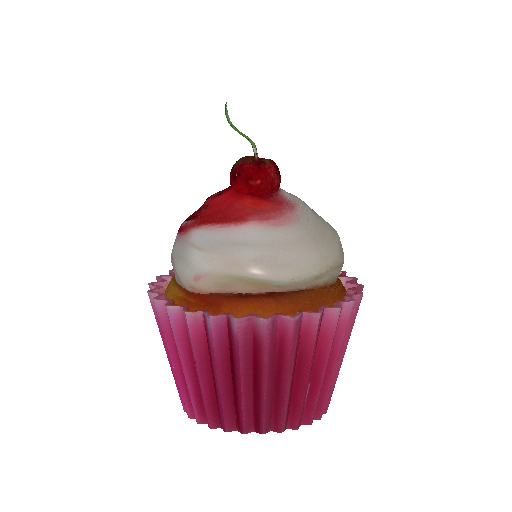} &
\interpfigte{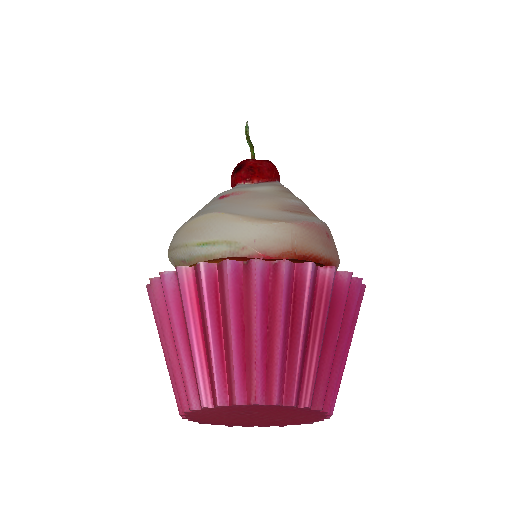} &
\interpfigte{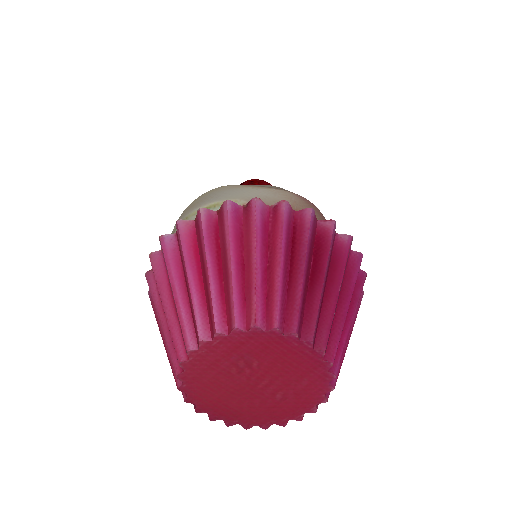} &
\\
\multicolumn{5}{c}{{Chocolate Ganache Cupcake}} & 
\multicolumn{5}{c}{{Strawberry Shortcake Cupcake}}
\\

\interpfigte{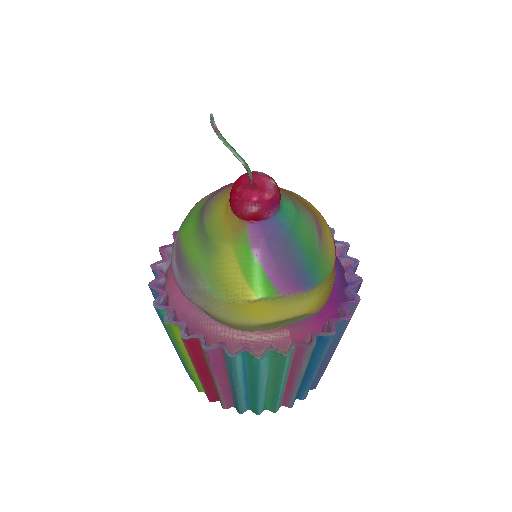} &
\interpfigte{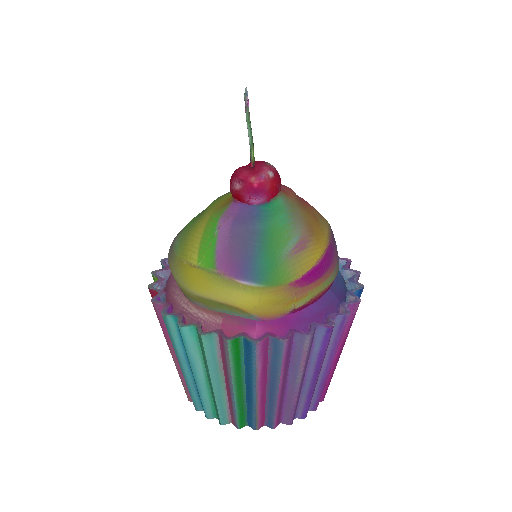} &
\interpfigte{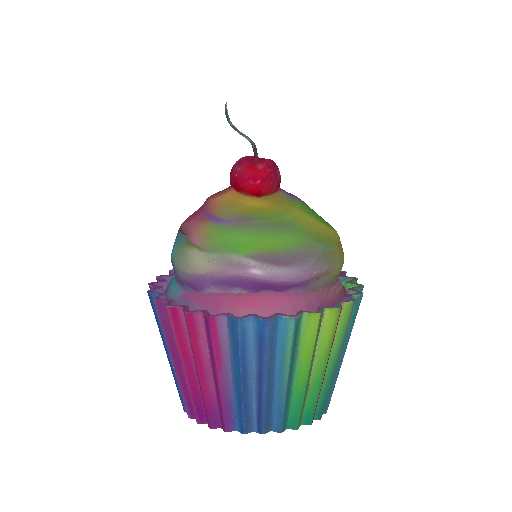} &
\interpfigte{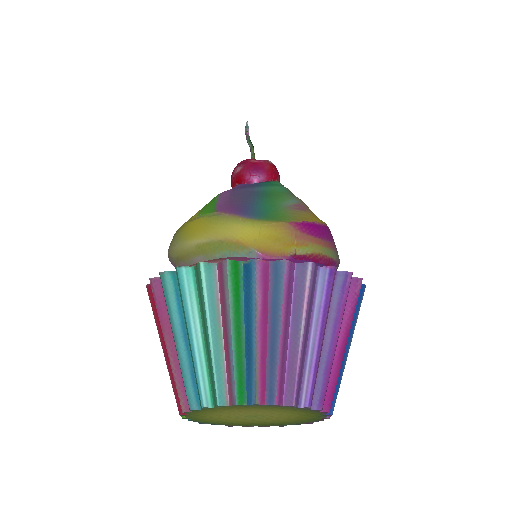} &
\interpfigte{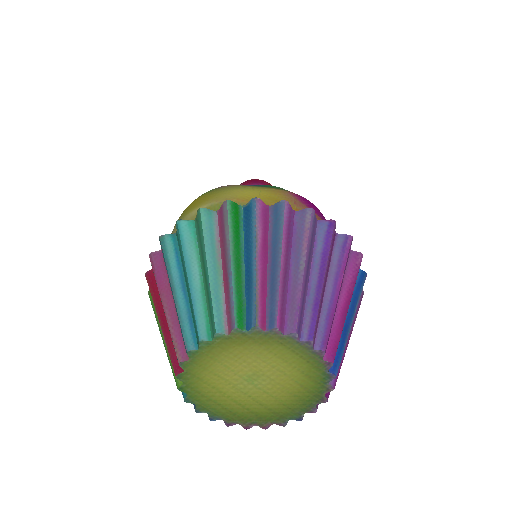} &
\interpfigte{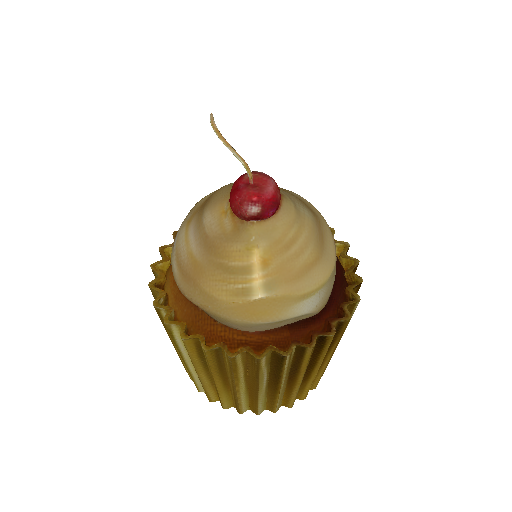} &
\interpfigte{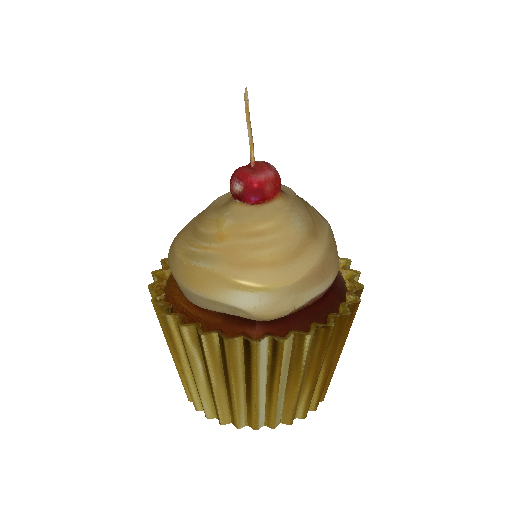} &
\interpfigte{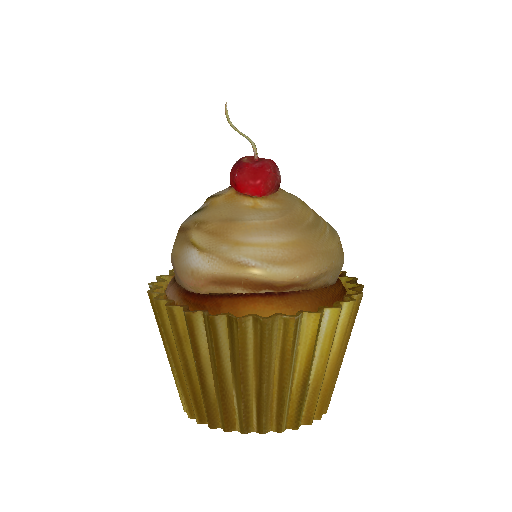} &
\interpfigte{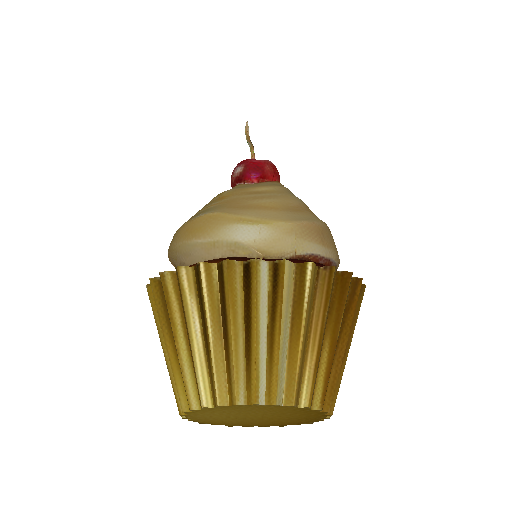} &
\interpfigte{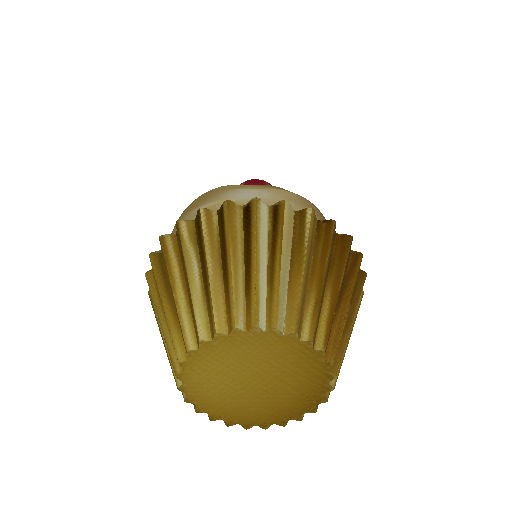} &
\\
\multicolumn{5}{c}{{Shiny Cupcake with Rainbow Colors}} & 
\multicolumn{5}{c}{{Gold Leaf Cupcake}}

\end{tabular}
\caption{Qualitative results of our method with various text prompts demonstrate that it generates diverse and multi-view consistent outputs that align with the given prompt. }
\label{fig:results_ours_multiprompt}
\end{figure*}

\newcommand{\interpfigtips}[1]{\includegraphics[trim=0 0cm 0cm 0cm, clip, width=1.6cm]{#1}\centering}
\newcommand{\interpfigtipp}[1]{\includegraphics[trim=2cm 0.5cm 2cm 0.5cm, clip, width=1.4cm]{#1}\centering}
\newcommand{\interpfigtippbb}[1]{\includegraphics[trim=3.5cm 0.5cm 3.5cm 0.5cm, clip, width=1.4cm]{#1}\centering}
\begin{figure*}[t]
\centering
\addtolength{\tabcolsep}{-1.3pt}   
\begin{tabular}{ccccccccccccccccc}
\\

\rotatebox{90}{Style Image} &
\multicolumn{2}{c}{\interpfigtips{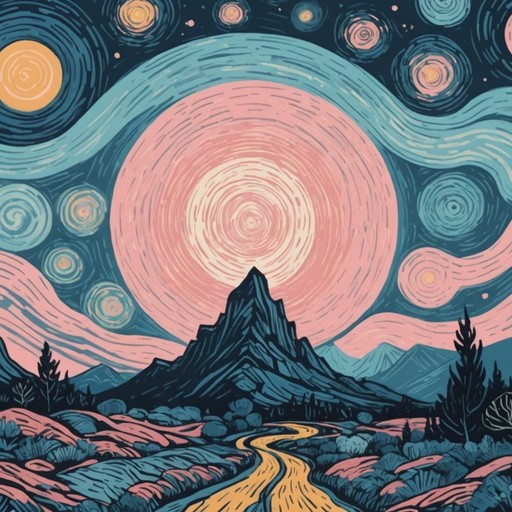}} &
\multicolumn{2}{c}{\interpfigtips{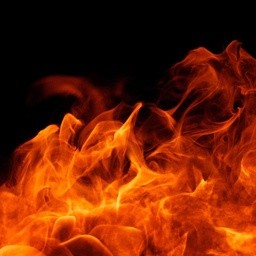}} &
\multicolumn{2}{c}{\interpfigtips{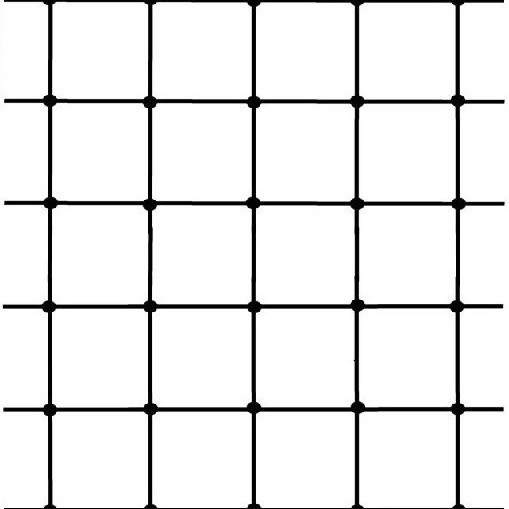}} &
\multicolumn{2}{c}{\interpfigtips{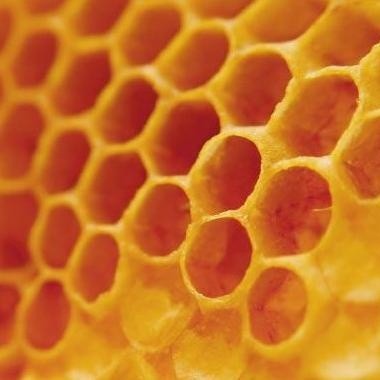}} &
\multicolumn{2}{c}{\interpfigtips{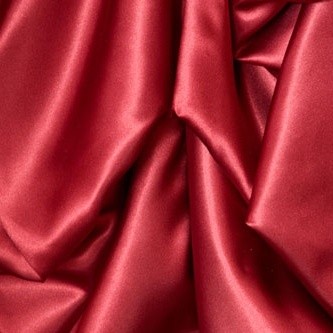}} &
\\
\rotatebox{90}{~~~~~Bag} &
\interpfigtippbb{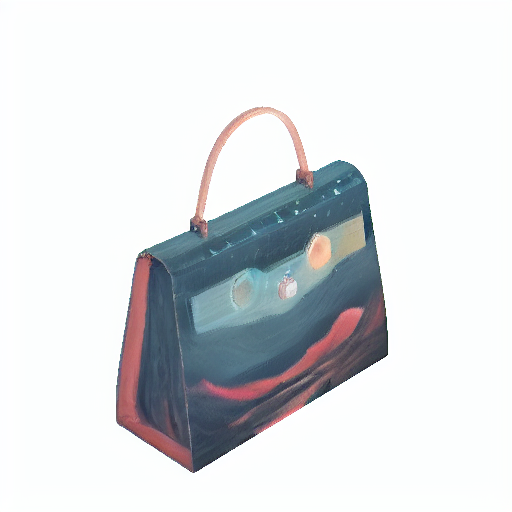} &
\interpfigtippbb{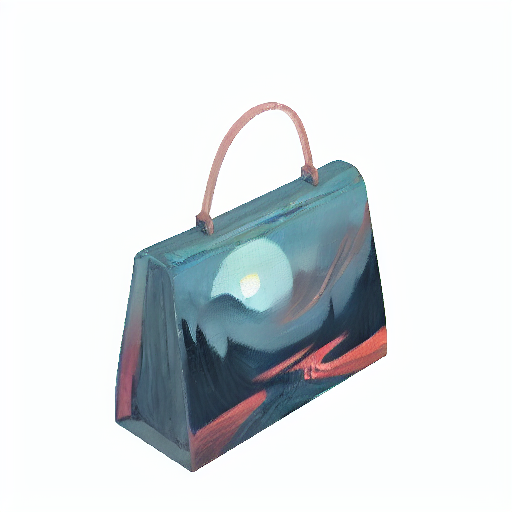} &
\interpfigtippbb{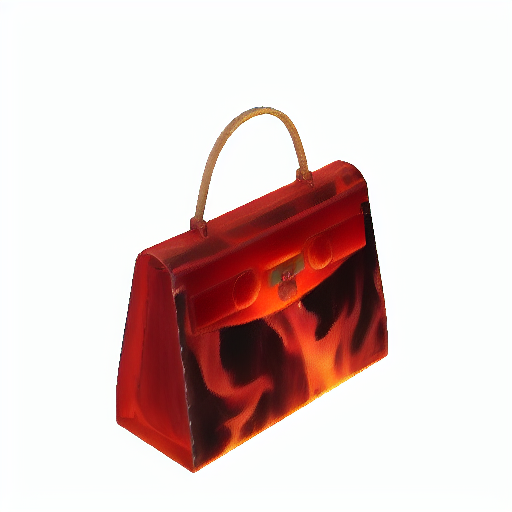} &
\interpfigtippbb{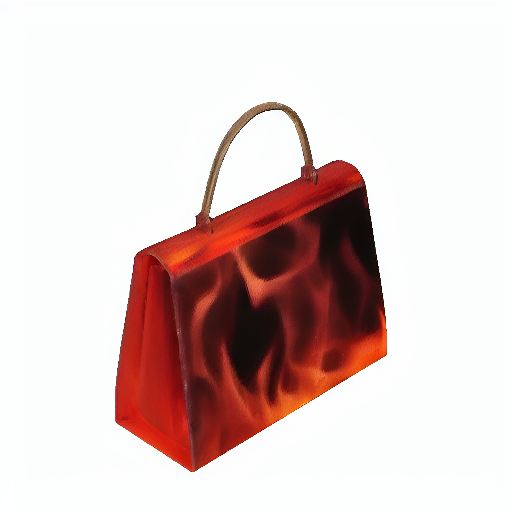} &
\interpfigtippbb{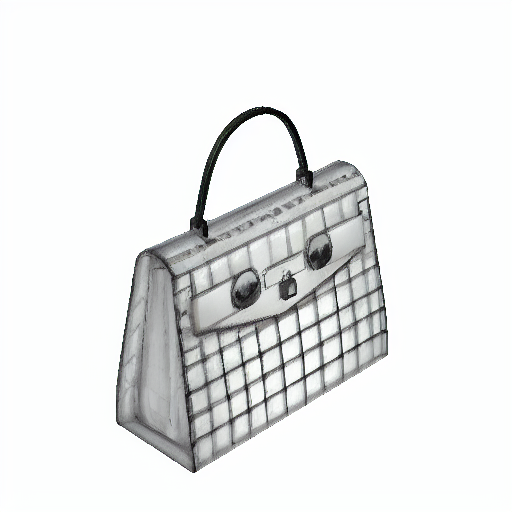} &
\interpfigtippbb{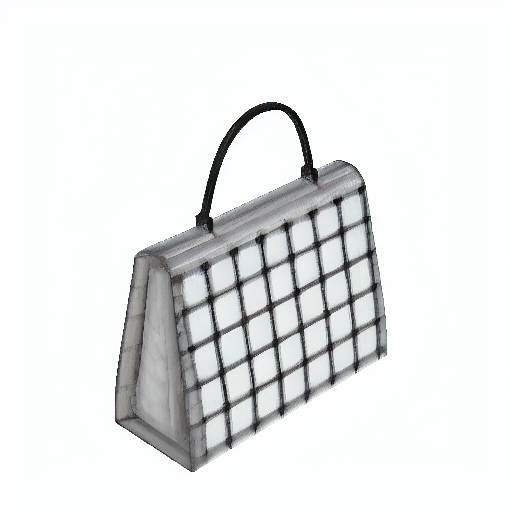} &
\interpfigtippbb{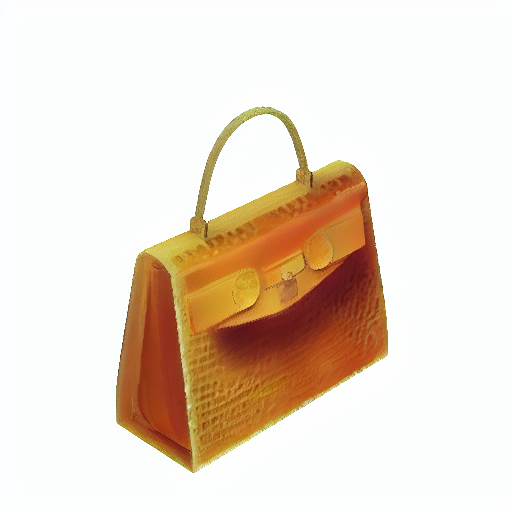} &
\interpfigtippbb{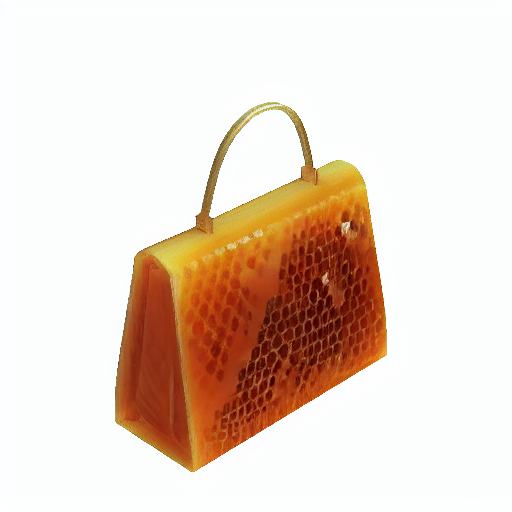} &
\interpfigtippbb{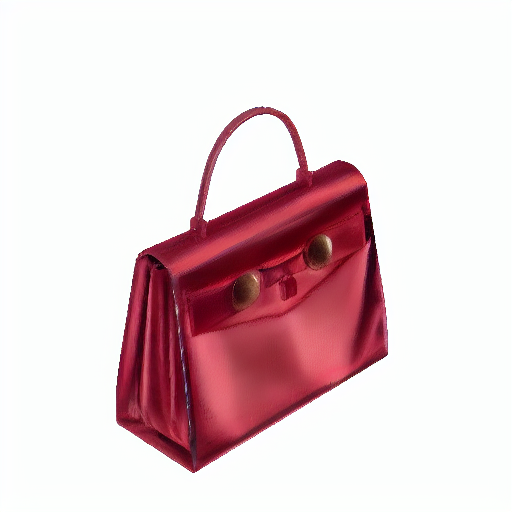} &
\interpfigtippbb{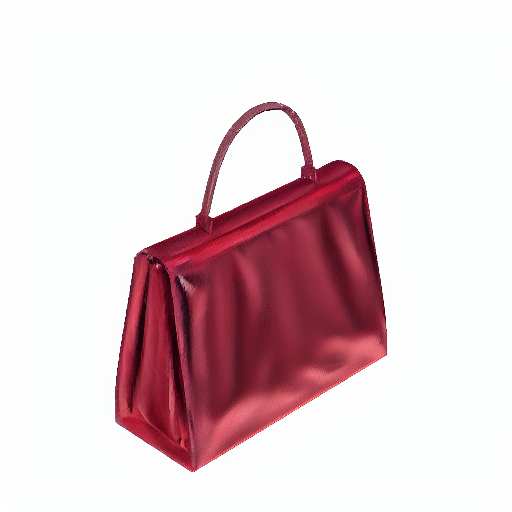} &
\\
\rotatebox{90}{Birdhouse} &
\interpfigtipp{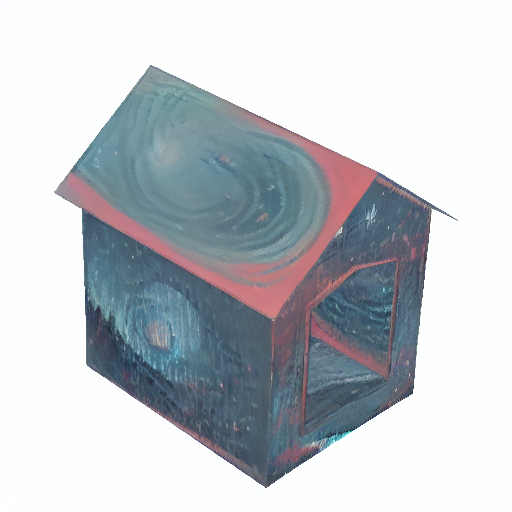} &
\interpfigtipp{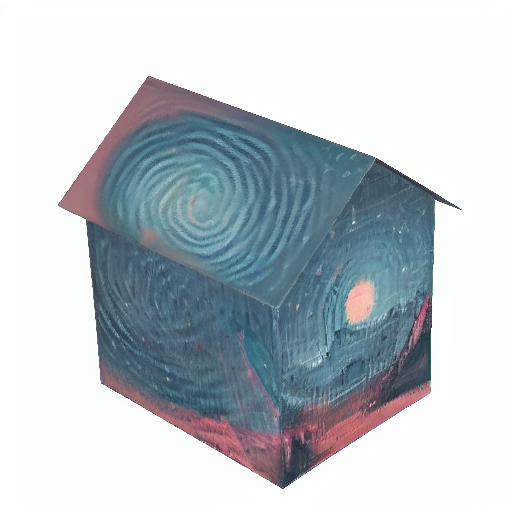} &
\interpfigtipp{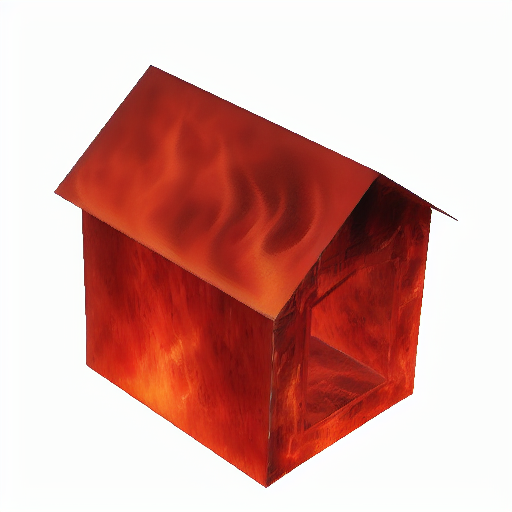} &
\interpfigtipp{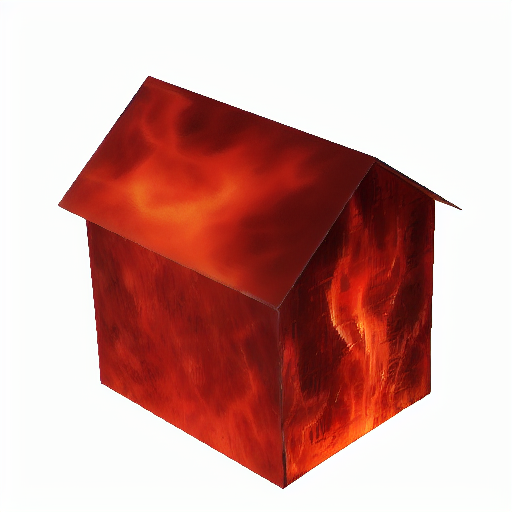} &
\interpfigtipp{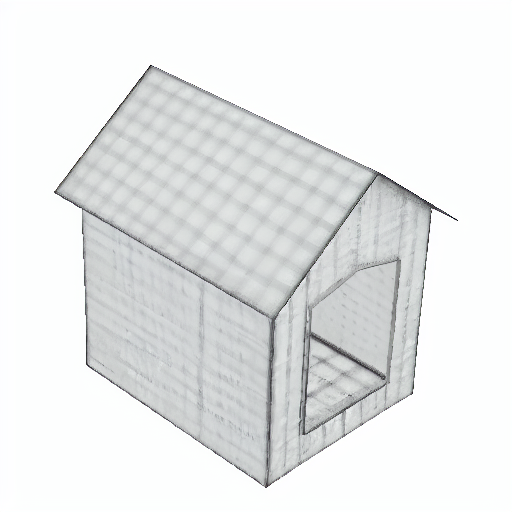} &
\interpfigtipp{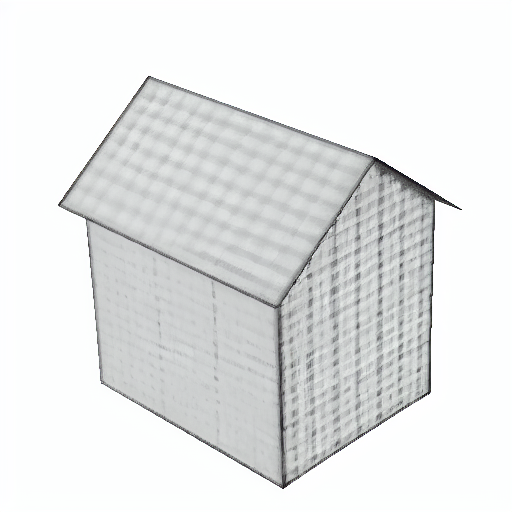} &
\interpfigtipp{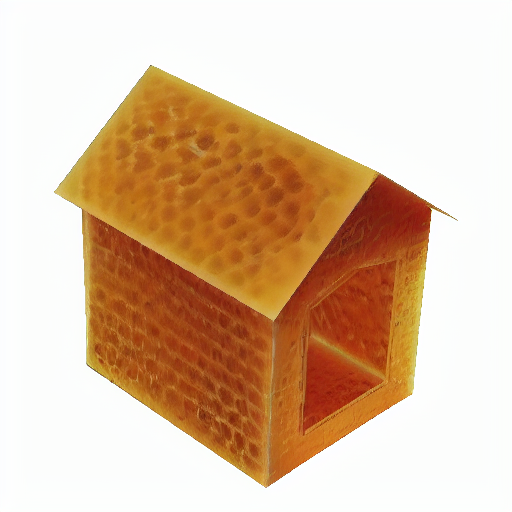} &
\interpfigtipp{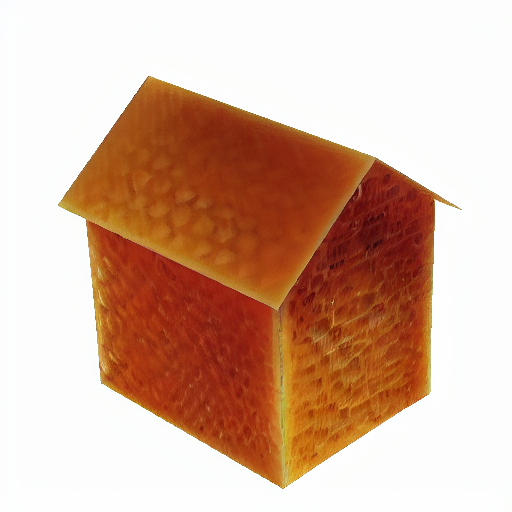} &
\interpfigtipp{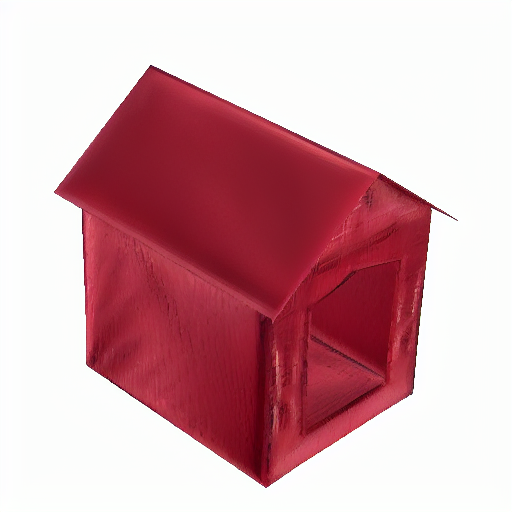} &
\interpfigtipp{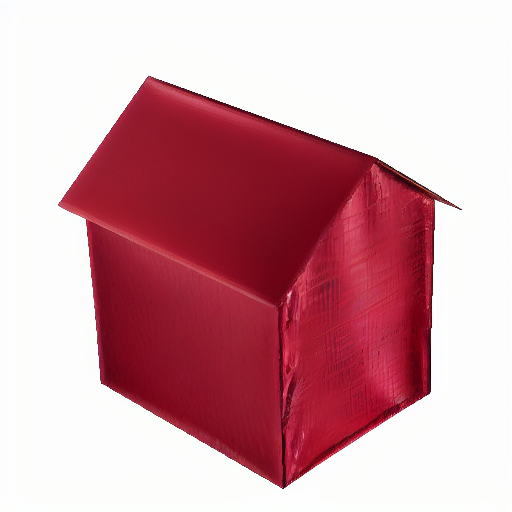} &
\\
\end{tabular}
\caption{Qualitative results of our method with IP-Adapter \cite{ye2023ip} guidance. Style image refers to the IP-Adapter input that contains the style to be transferred, which guides the texture generation process. Front and back view rendering of the generated textures are provided in separate rows.}
\label{fig:results_ip_adapter}
\end{figure*}

\textbf{Analysis and Ablation Study.} To better understand the texture generation process, we visualize intermediate outputs at different timesteps of the denoising process in Fig. \ref{fig:texture_progressing}. While the original diffusion process spans 1000 timesteps, we use 20-step DDIM sampling for more efficient generation. The process begins by predicting $x_0$ from $x_t$ at $t=1000$, then predicting $x_0$ from $t=950$, and so on. We visualize the predictions from these intermediate steps in Fig. \ref{fig:texture_progressing}.
In total, only 70\% of the timesteps are applied, which corresponds to a final timestep of 300 in the original diffusion scale. At this point, texture generation converges, and no further improvements are observed. At each sampled timestep, we extract the predicted clean image $x_0$ from multiple views and map them to the shared UV texture space. This enables us to observe how the predicted textures evolve during the denoising process.
In the early stages (i.e., at higher timesteps), the $x_0$ predictions mainly capture coarse structures and lack fine details. As denoising progresses and the timestep decreases, the model refines the textures, progressively introducing sharper features and more accurate details. 

In Fig. \ref{fig:results_sablation} and Table \ref{table:abl}, we present a qualitative and quantitative ablation study. Through our experiments, we have observed that the challenge in texture generation lies in synthesizing intricate details. The multi-diffusion approach can often wash out many of these details in a straightforward manner.

\newcommand{\interpfigtexx}[1]{\includegraphics[trim=0cm 0cm 0cm 0cm, clip, width=1.5cm]{#1}}
\newcommand{\interpfigrenn}[1]{\includegraphics[trim=3cm 1cm 3cm 3cm, clip, width=1.5cm]{#1}}

\begin{figure}[t!]
\centering
\addtolength{\tabcolsep}{1pt}  
\begin{tabular}{cccccccc}
& \multicolumn{4}{c}{{A Ball with Nebula Pattern}}
\\
\rotatebox{90}{~~Texture} &
\interpfigtexx{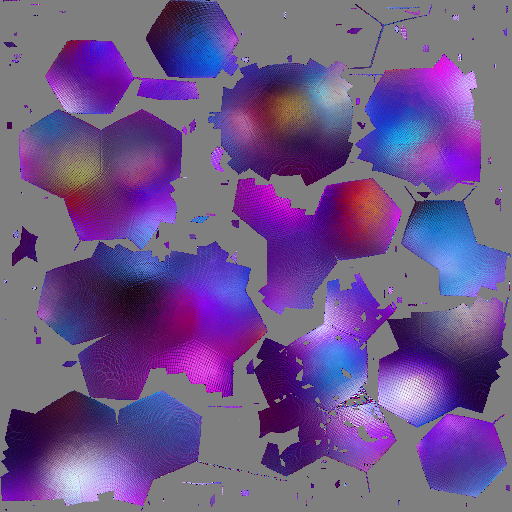} &
\interpfigtexx{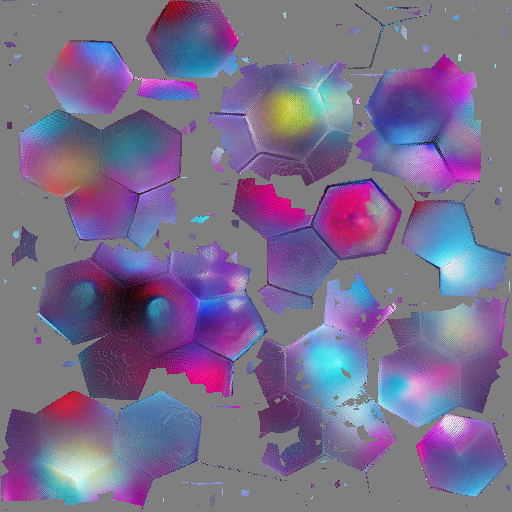} &
\interpfigtexx{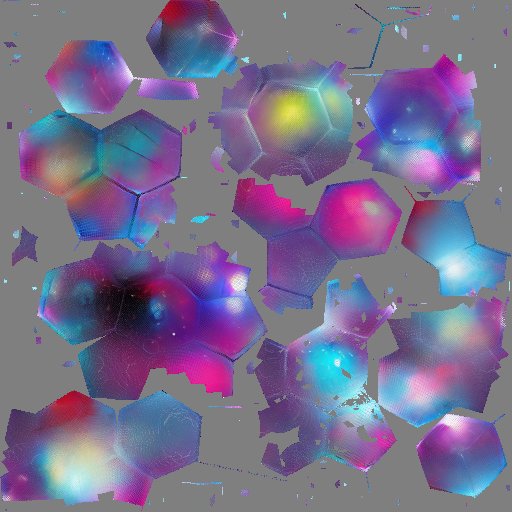} &
\interpfigtexx{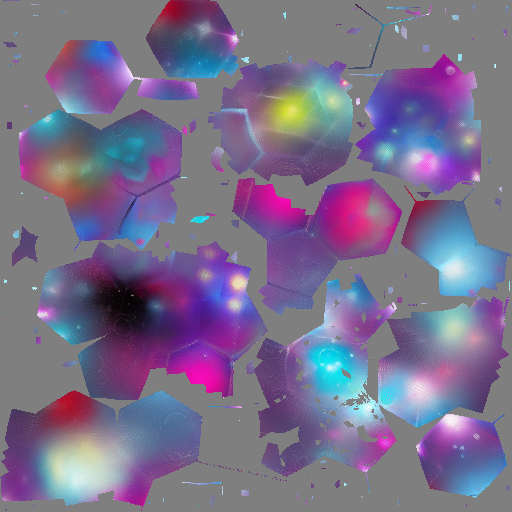} 
\\
\rotatebox{90}{~~~~Render} &
\interpfigrenn{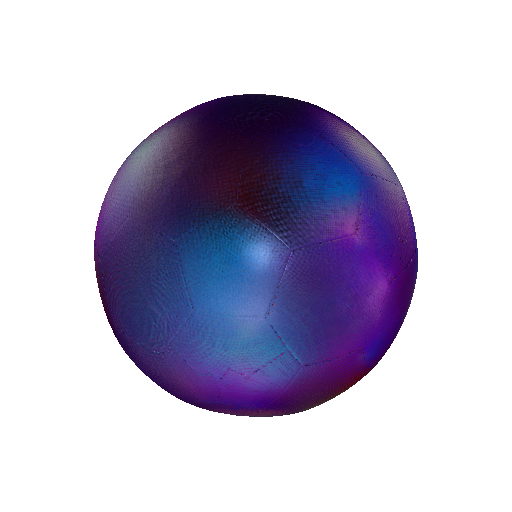} &
\interpfigrenn{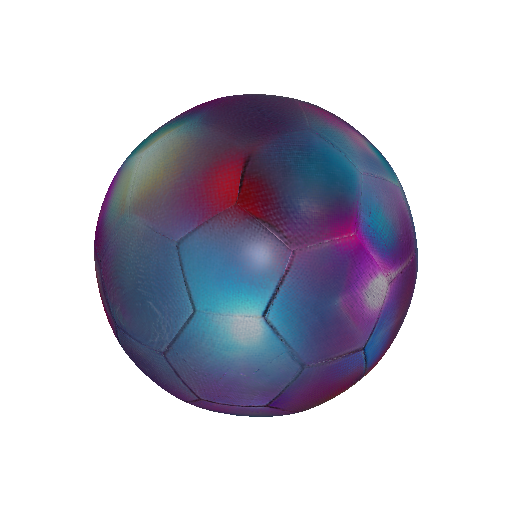} &
\interpfigrenn{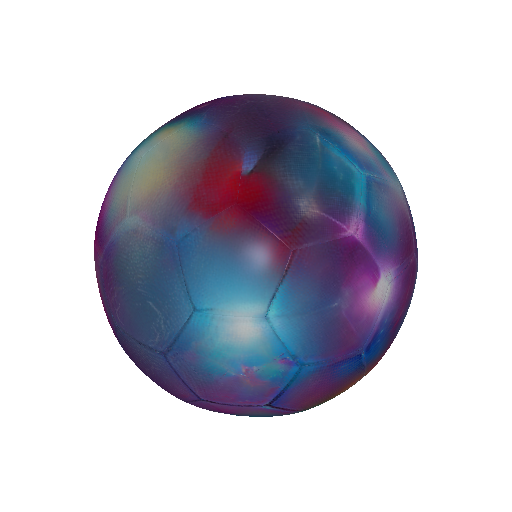} &
\interpfigrenn{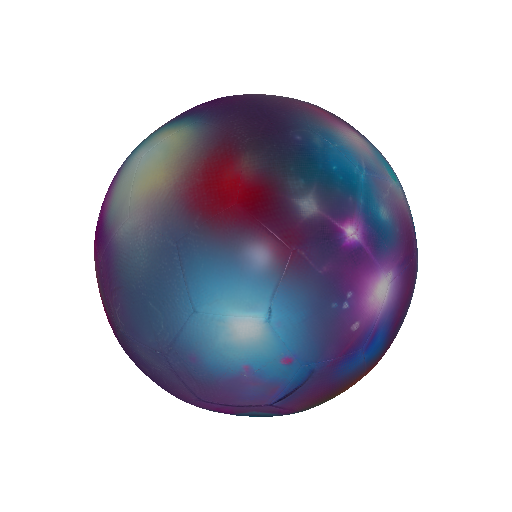} 
\\
& \multicolumn{4}{c}{{A Marble-Patterned Ball}} 
\\
\rotatebox{90}{~~Texture} &
\interpfigtexx{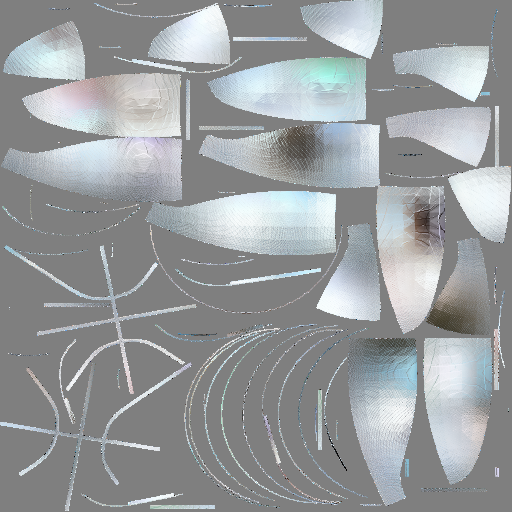} &
\interpfigtexx{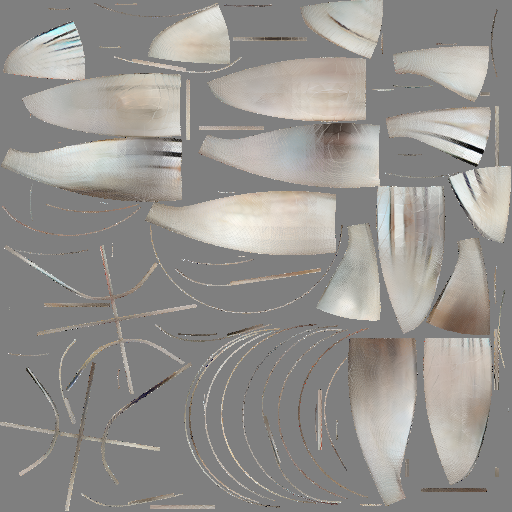} &
\interpfigtexx{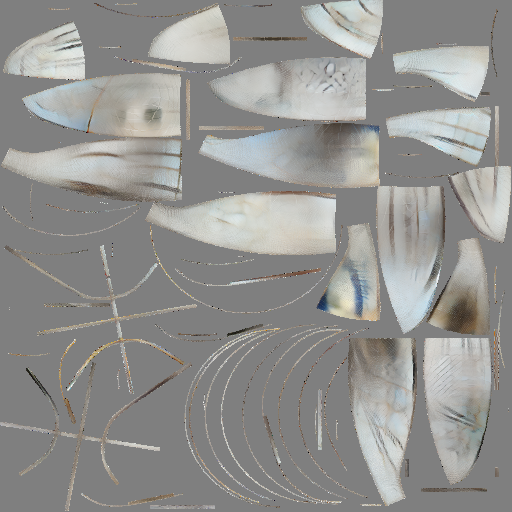} &
\interpfigtexx{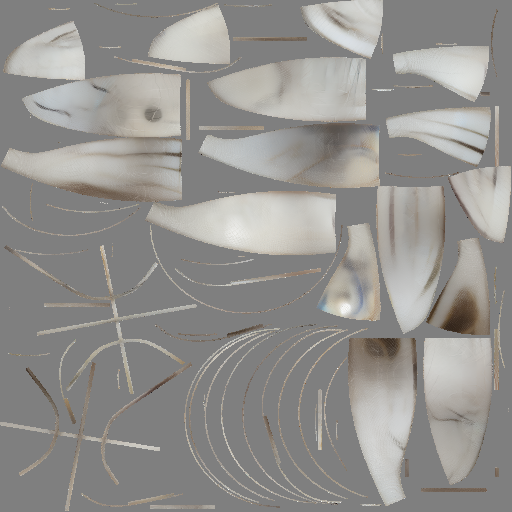}  
\\
\rotatebox{90}{~~~~Render} &
\interpfigrenn{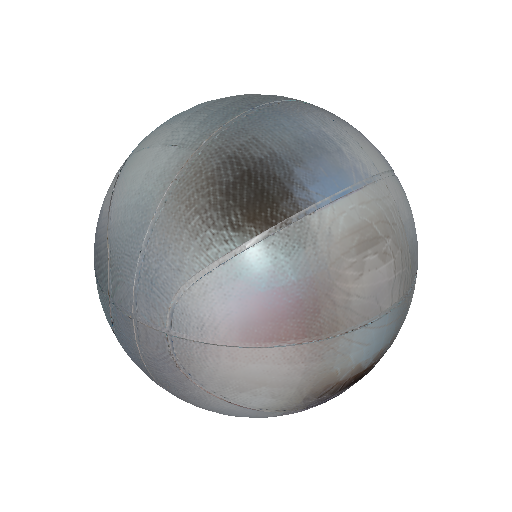} &
\interpfigrenn{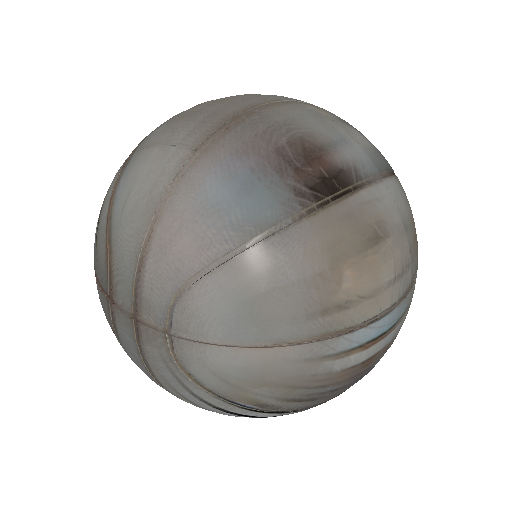} &
\interpfigrenn{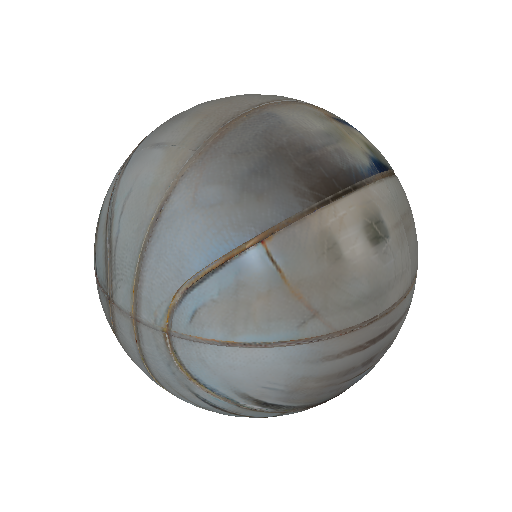} &
\interpfigrenn{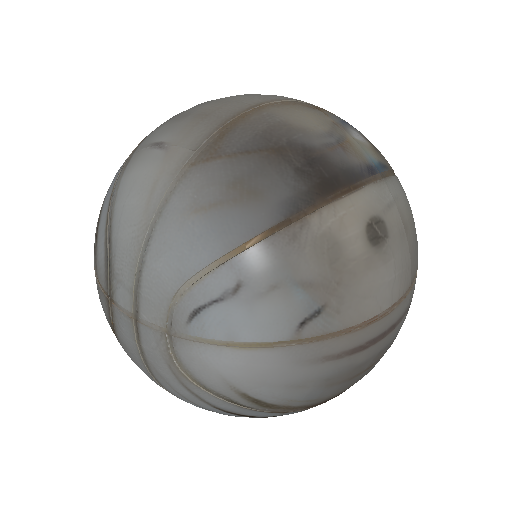} 
\\
& t=900
& t=700
& t=500
& t=300
\\
\end{tabular}

\caption{Texture ($x_0$) and image generations at different timesteps ($t$) during the denoising process. Textures are obtained by mapping each view’s $x_0$ prediction to the UV space, using denoising results at various timesteps $t$. As denoising progresses and the timestep decreases, finer details progressively appear. Generation continues until 70\% of the denoising process is complete, ending at timestep 300, at which point texture generation converges, and no further improvements are observed.}

\label{fig:texture_progressing}
\end{figure}

\newcommand{\interpfigtab}[1]{\includegraphics[trim=3cm 2.5cm 3cm 2.5cm, clip, width=3cm]{#1}}
\begin{figure}[h]
\centering
\addtolength{\tabcolsep}{1pt}
\resizebox{\linewidth}{!}{
\begin{tabular}{cccccccc}
\\
\interpfigtab{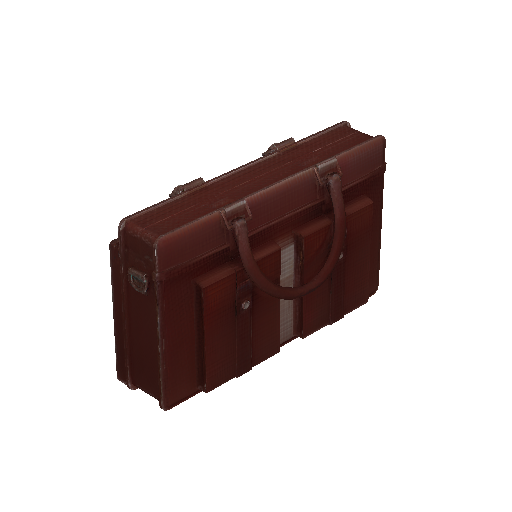} &
\interpfigtab{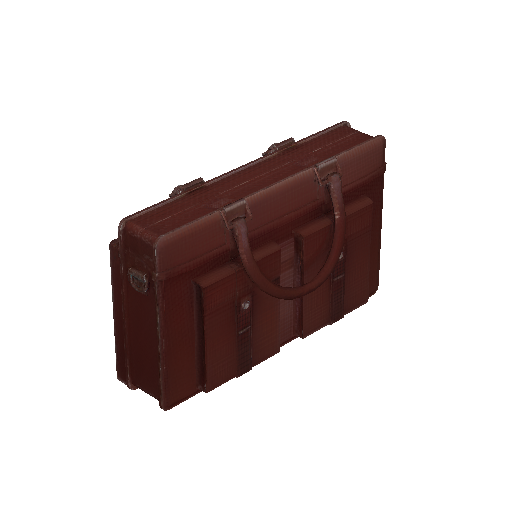} &
\interpfigtab{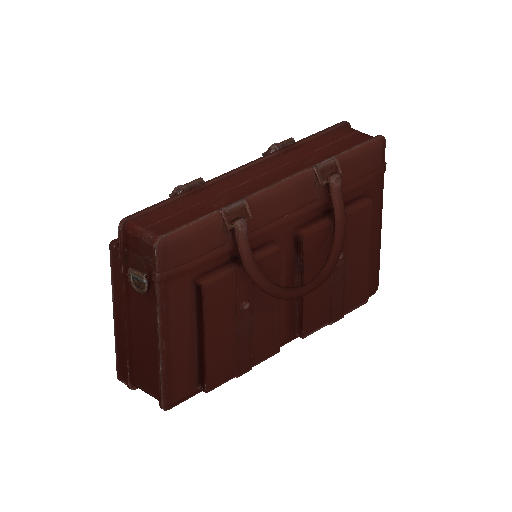} &
\interpfigtab{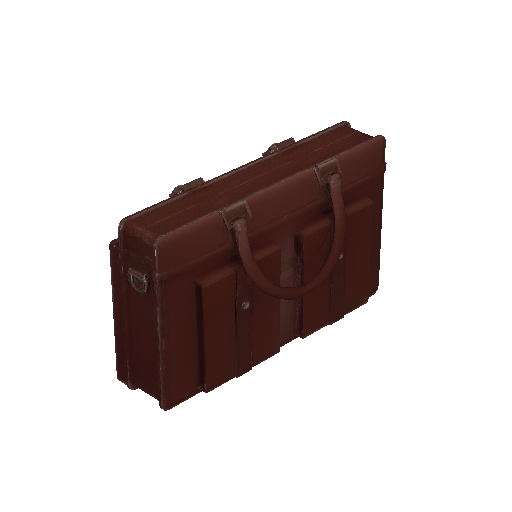} &
\interpfigtab{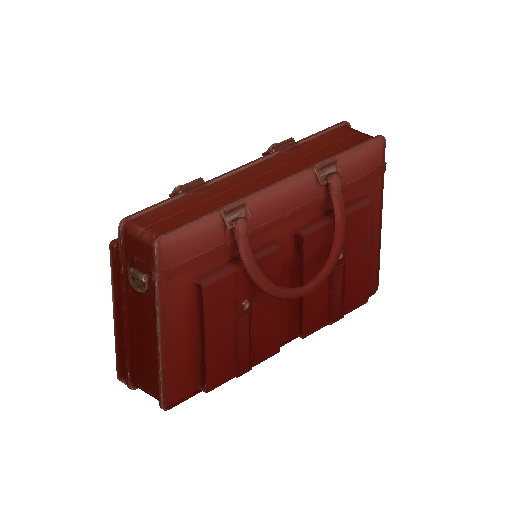} 
\\
\interpfigtab{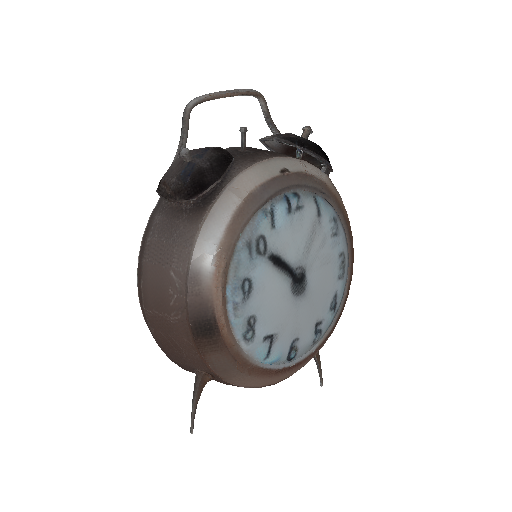} &
\interpfigtab{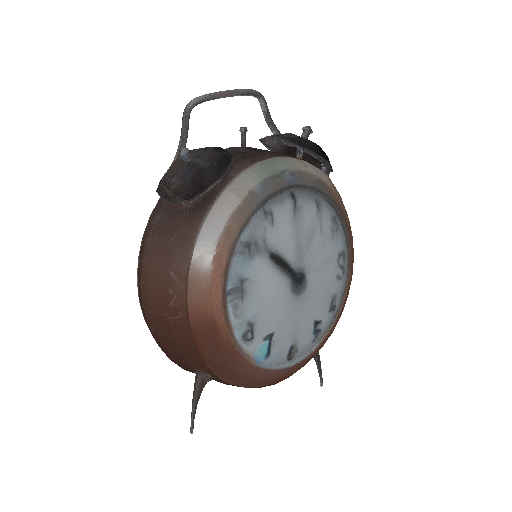} &
\interpfigtab{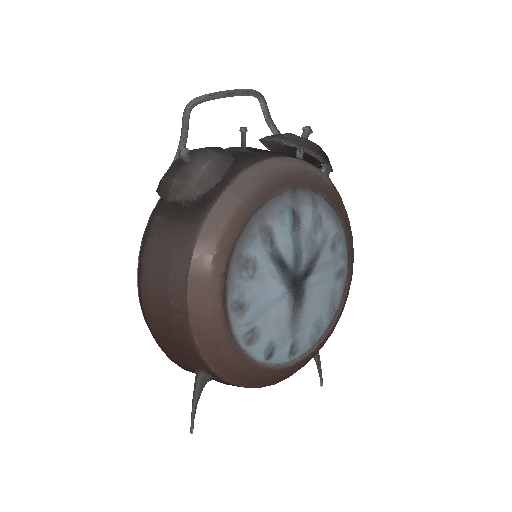} &
\interpfigtab{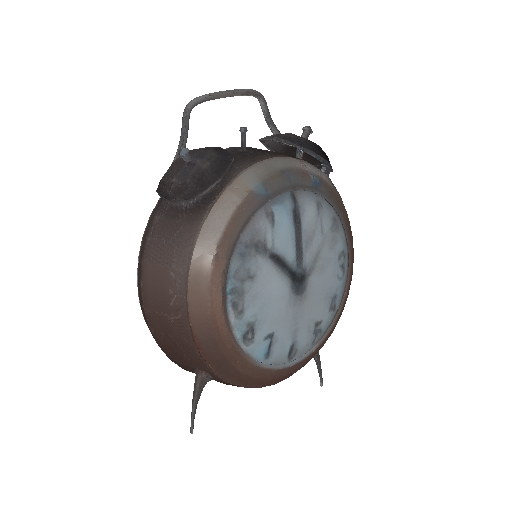} &
\interpfigtab{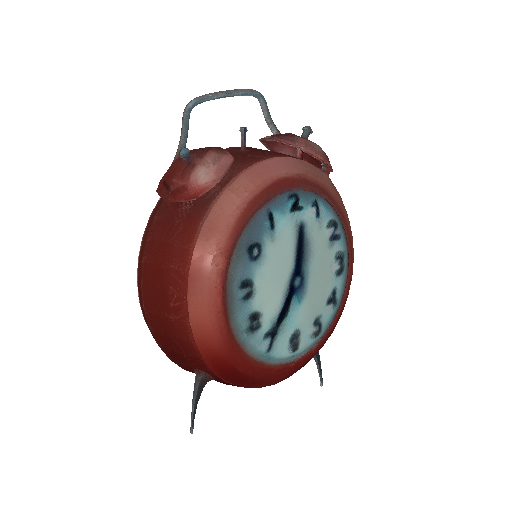} 
\\
\interpfigtab{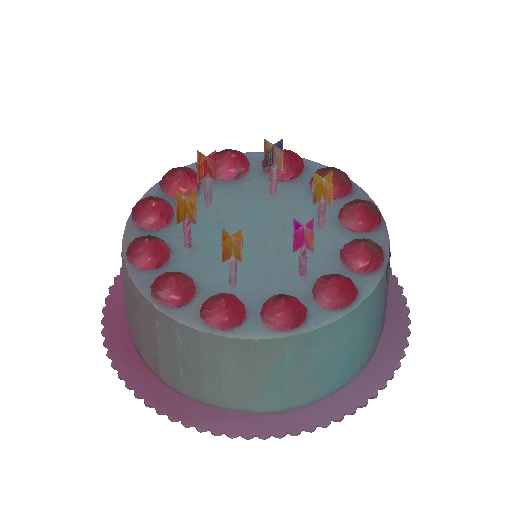} &
\interpfigtab{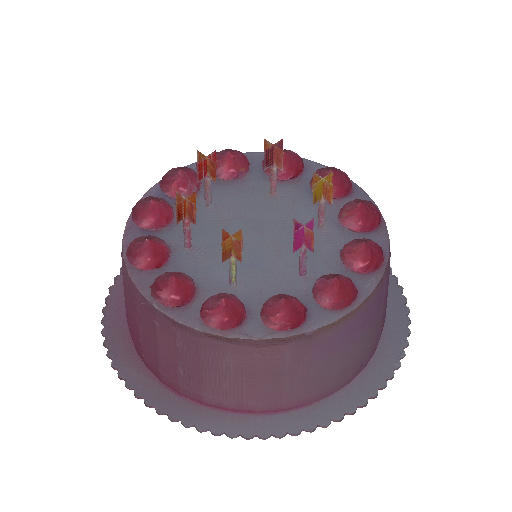} &
\interpfigtab{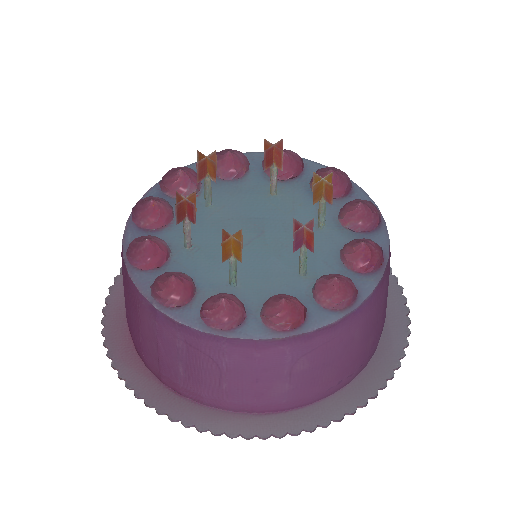} &
\interpfigtab{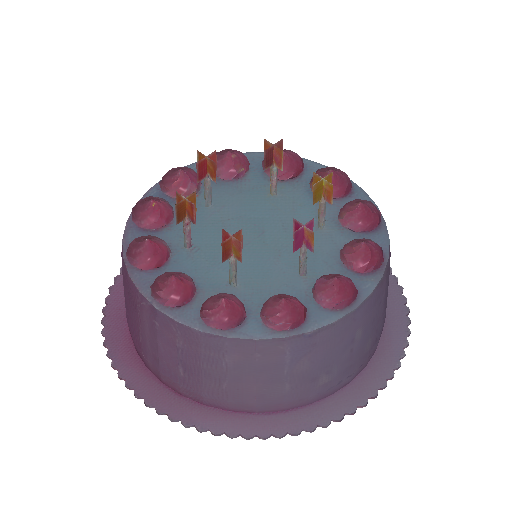} &
\interpfigtab{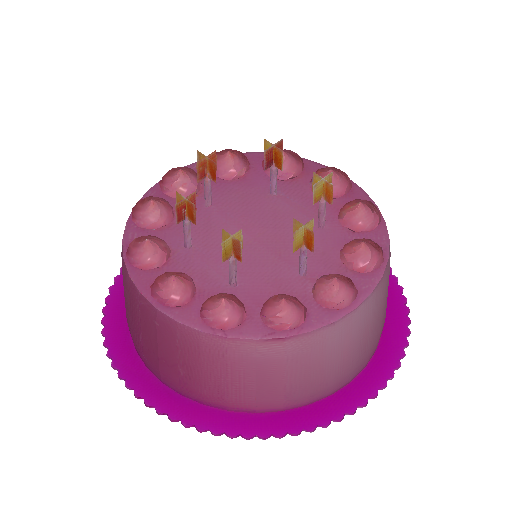} 
\\
\interpfigtab{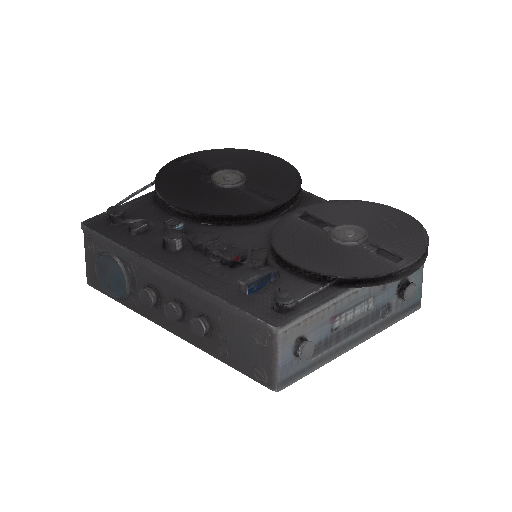} &
\interpfigtab{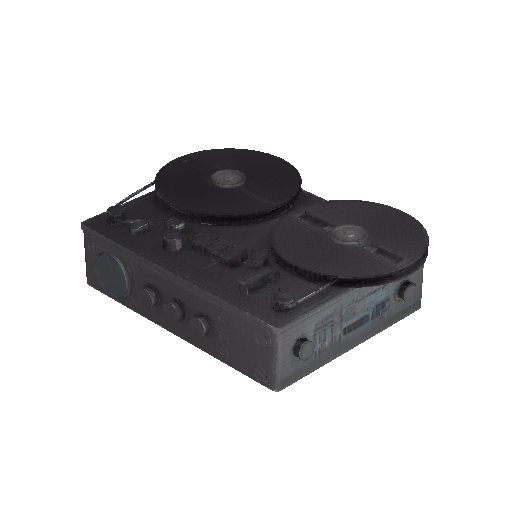} &
\interpfigtab{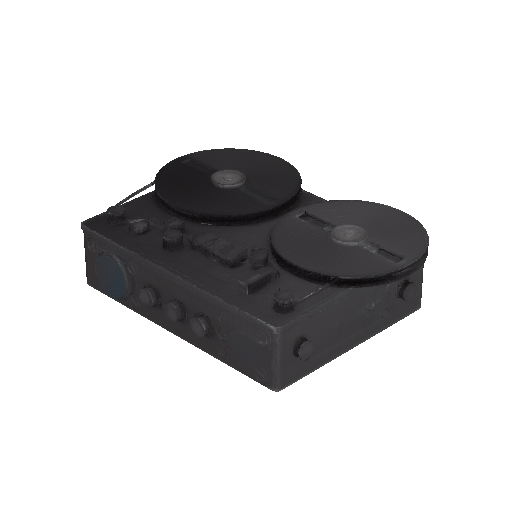} &
\interpfigtab{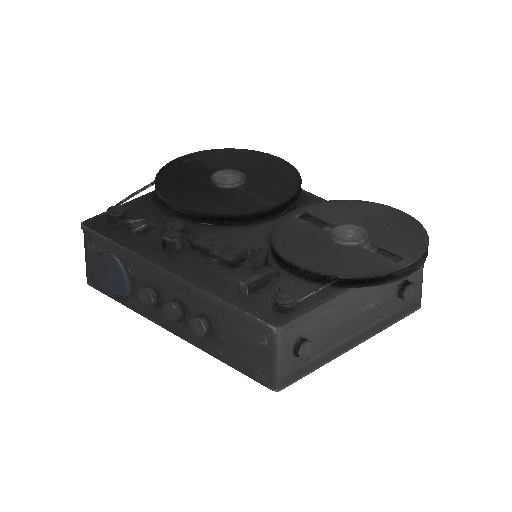} &
\interpfigtab{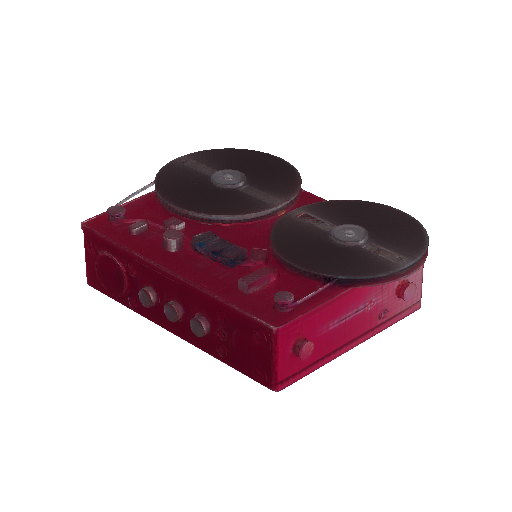} 
\\
\interpfigtab{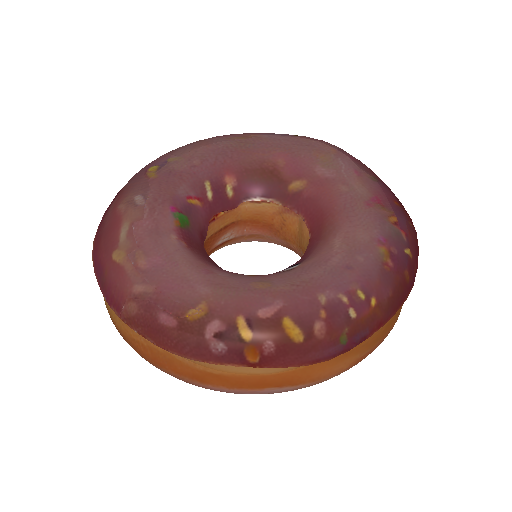} &
\interpfigtab{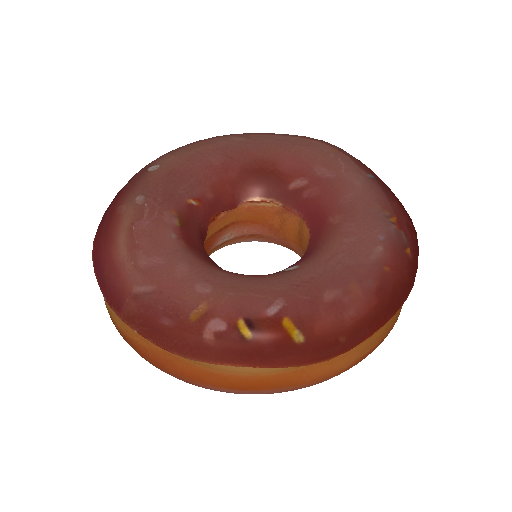} &
\interpfigtab{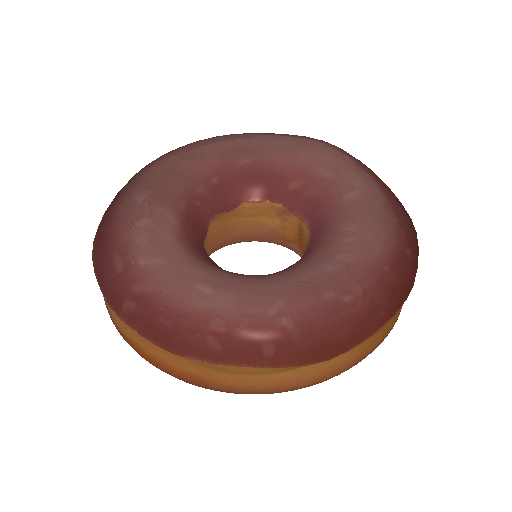} &
\interpfigtab{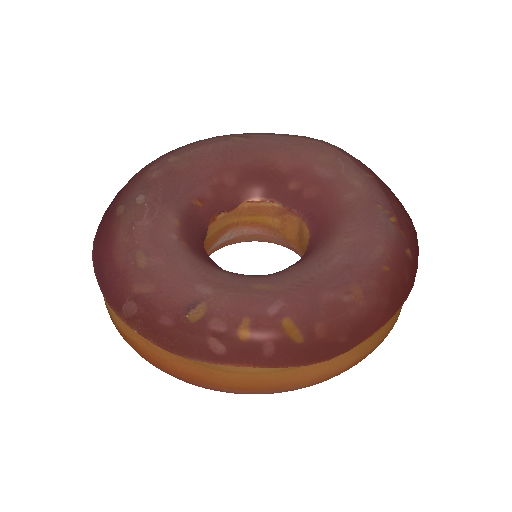} &
\interpfigtab{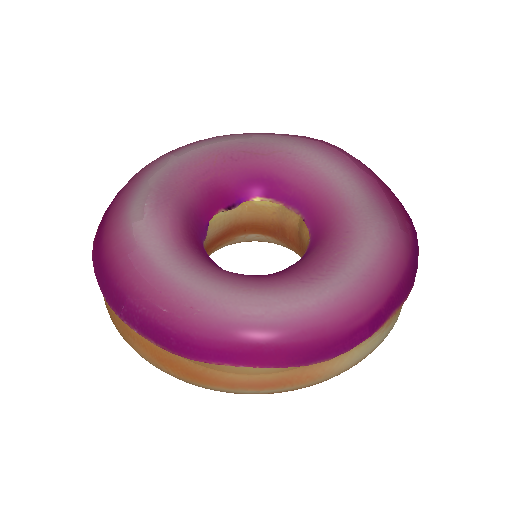} 
\\
Ours 
& w/o multi-scale  
& w/o norm 
& w/ norm avg. 
& w/o modified denoising
\\

\end{tabular}
}
\caption{Qualitative results of our Ablation Study. The main challenge of generating textures for shapes in our experiments was to synthesize realistic details. As can be seen, our contributions collectively yield realistic textures. }
\label{fig:results_sablation}
\end{figure}

Firstly, we showcase the results of our final model, which incorporates the modified denoising, multi-scale generations, and softmax-based norm-guided averaging in Fig. \ref{fig:results_sablation}. Next, we compare it with a version where multi-scale generation is omitted, and the textures are consistently projected to a $512\times512$ resolution. As evident, without multi-scale generation, the image details appear less sharp.
This is primarily caused by disagreements in the generations projected to different pixels in the UV map. These projections lack guidance from the multi-diffusion step early on due to the sparse projection to the UV maps. This lack of guidance leads to divergent generations when the coarse shape is established, resulting in different views attempting to generate various image details in later steps. Consequently, the model struggles to preserve these details as they are canceled out via the averaging of overlapping pixels.
Contrary to our visual inspection, we note that the FID of the model with and without multi-scale is similar. Consequently, we conducted a user study for this setting as well. Users preferred the results of the multi-scale model by 85\%.

\begin{table}
\caption{Ablation Study of View Merging.
These experiments use 36 different views.}
\centering
\label{table:abl}
\begin{tabular}{|l|c|c|}
\hline
Method & FID & KID ($\times 10^{-3}$)  \\

\hline
w/o multi-scale & 16.60 & 1.79 \\
w/o norm & 19.03 & 2.56 \\
w norm avg. & 18.14 & 2.31 \\
w/o modified denoising & 18.23 & 3.14 \\
\hline
Ours & 16.63 & 1.73 \\
\hline
\end{tabular}
\end{table}

Next, we proceed to compare with alternative methods of incorporating normal map information in our multi-diffusion step. In these experiments, we reintroduce multi-scale generation. It is worth noting that our final model employs softmax-based weighting derived from the normal maps. This is because not every view is reliable for each pixel in the UV map. Instead of employing the softmax operation, we experiment by directly utilizing the normal weights in our weighted multi-diffusion step, as well as by completely omitting the use of normal weights.
As shown in Fig. \ref{fig:results_sablation}, using softmax-based weighting helps us achieve sharper results compared to those two settings. 
We also observe consistent FID improvements from not using the normal map information at all to using them via direct averaging and finally using them via softmax averaging in Table \ref{table:abl}.

Subsequently, we present the outcomes when the proposed denoising strategy, referred to as \emph{Modified Denoising}, is omitted. Previously, we displayed these results for the Stable Diffusion method, showcasing single-image generations without any multi-diffusion for the textures in Fig. \ref{fig:method_noising}. 
In that case, the colors became oversaturated, and within our framework, it also led to object texturing with uniformly bright colors, as demonstrated in Fig. \ref{fig:results_sablation}.
This compromises the realism and diminishes the effectiveness of prompt-based control. For instance, both the CD player and the cake exhibit the same coloring for distinct parts.

\begin{figure}[t]
\centering
\includegraphics[width=1.0\linewidth]{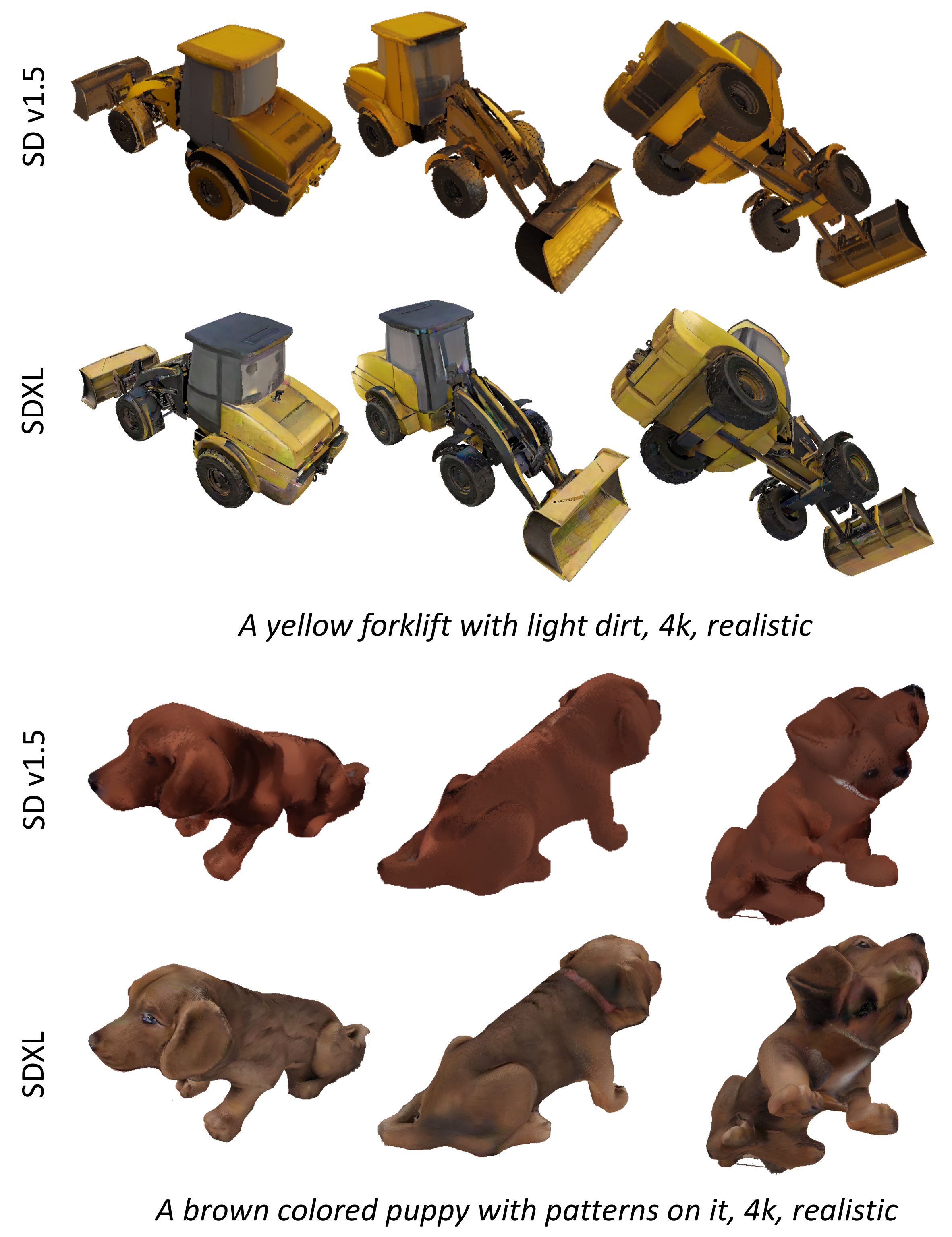}
\caption{Visual results with different SD base models.}
\label{fig:sd_baseline_comp}
\end{figure}

Next, we present the results of our camera selection method in Fig. \ref{fig:results_camera}, demonstrating that the proposed view selection yields higher-quality textures compared to uniformly sampled camera parameters for the same number of views. It is important to note that in our previous comparisons, we used the same fixed views to ensure a fair evaluation.
Lastly, our method is not restricted to a single Stable Diffusion model. While our previous experiments utilized SDv1.5, Fig. \ref{fig:sd_baseline_comp} presents results obtained using SDXL.

\newcommand{\interpfigmmabl}[1]{\includegraphics[trim=2.5cm 1cm 2cm 1cm, clip, width=2.4cm]{#1}}
\begin{figure}[t]
\scalebox{0.8}{
\centering  
\begin{tabular}{cccccc}

\rotatebox{90}{~~~Fixed Views} &
\interpfigmmabl{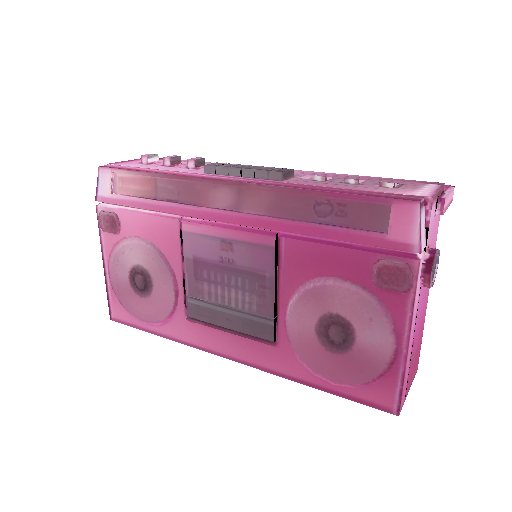} &
\interpfigmmabl{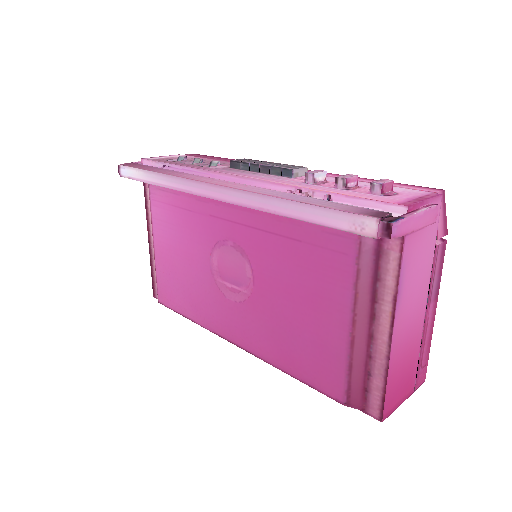} & 
\interpfigmmabl{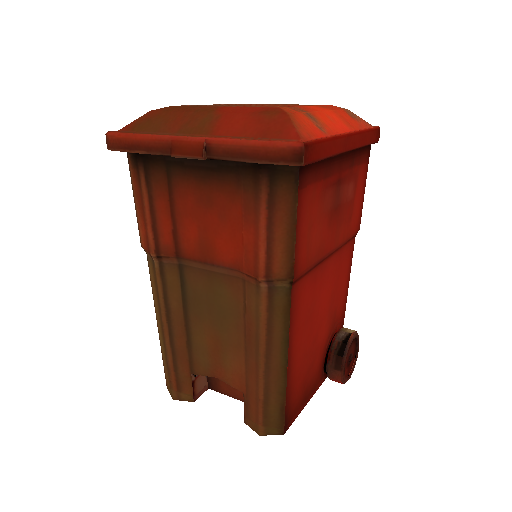} &
\interpfigmmabl{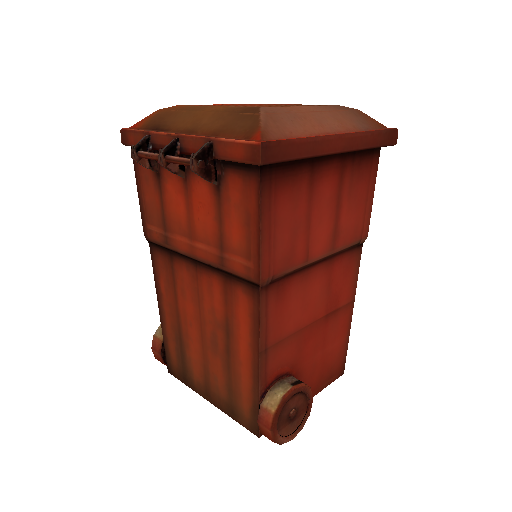} 
\\
\rotatebox{90}{~~Selected Views} &
\interpfigmmabl{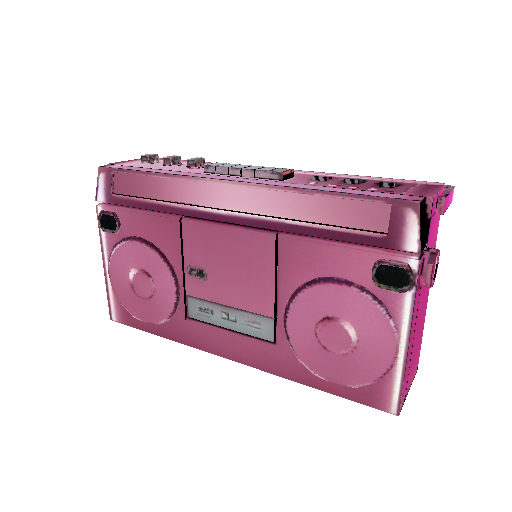} &
\interpfigmmabl{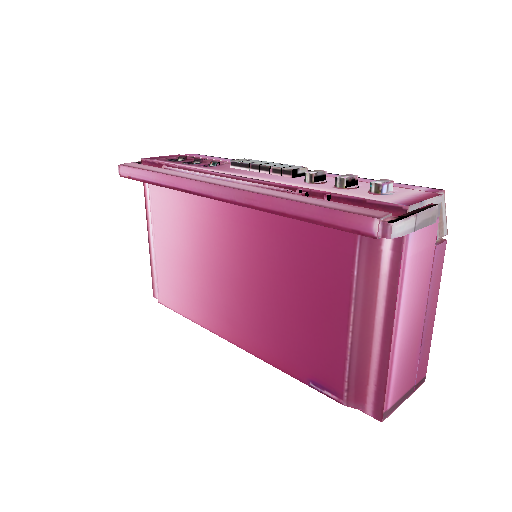} & 
\interpfigmmabl{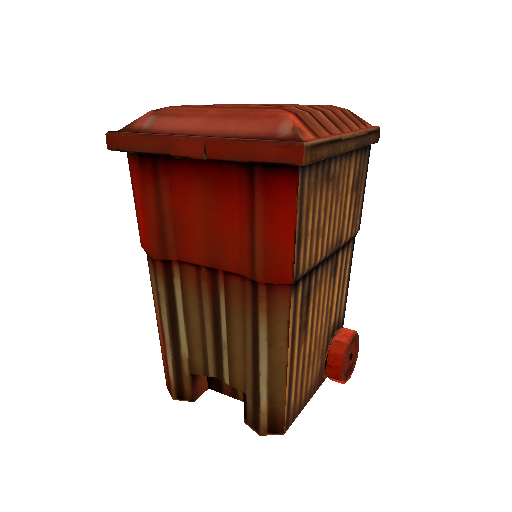} &
\interpfigmmabl{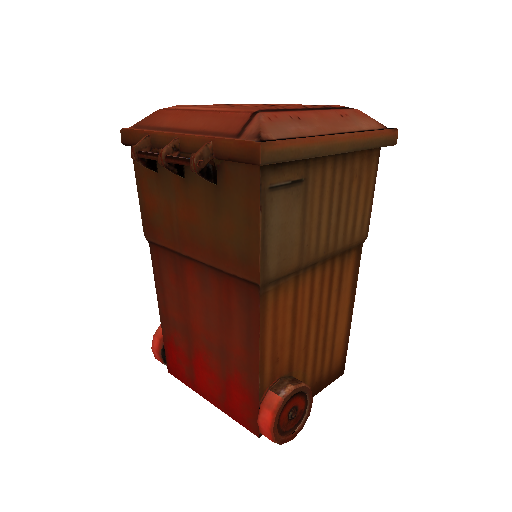} 
\\
&
\multicolumn{2}{c}{{A pink metallic}} &
\multicolumn{2}{c}{{A dumpster with rugged}} &

\\
&
\multicolumn{2}{c}{{CD player}} &
\multicolumn{2}{c}{{rust textures}} &
\\
\end{tabular}
}
\caption{With automatic view selection, we observe higher quality texture generations compared to uniformly sampling fixed views. 16 views are used in this experiment for both settings. }
\label{fig:results_camera}
\end{figure}

\section{Conclusion}

In this paper, we introduce MD-ProjTex, a novel pipeline that leverages the Multi-Diffusion technique in uv-space for text-prompted texture generation. By integrating denoising directions obtained from reference models across different views, our algorithm ensures consistency during generation, leading to significantly faster processing speeds compared to sequential methods. Our contributions include the proposal of a training-free pipeline for text-prompted texture generation, parallel denoising of multiple views, and extensive experimental validation demonstrating the effectiveness of our framework. Through quantitative and qualitative analyses, we show that MD-ProjTex outperforms state-of-the-art methods in terms of both speed and quality, offering a promising solution for high-quality 3D content generation.

\bibliographystyle{IEEEtran}
\bibliography{egbib}

\end{document}